\PassOptionsToPackage{numbers,sort&compress,square}{natbib}

\documentclass{article}
\usepackage[preprint]{neurips_2025}

\usepackage[numbers,sort&compress,square]{natbib}
\setcitestyle{square,numbers,sort&compress}

\usepackage[hidelinks]{hyperref}

\usepackage{graphicx}
\usepackage{booktabs}
\usepackage{amssymb}
\usepackage{amsmath}
\usepackage{multirow}
\usepackage{multicol}
\usepackage{subcaption}
\usepackage{makecell}
\usepackage{pgffor}
\usepackage[T1]{fontenc}
\usepackage{siunitx}
\usepackage{needspace}
\usepackage{caption}
\usepackage{placeins}
\usepackage{textcomp}
\usepackage{bm}
\usepackage{mdframed}
\newmdenv[
  linecolor=black,
  linewidth=0.5pt,
  roundcorner=4pt,
  skipabove=6pt,
  skipbelow=6pt
]{examplebox}

  {\begin{examplebox}\textbf{#1.} }%
  {\end{examplebox}}

\title{PoETa v2: Toward More Robust Evaluation of Large Language Models in Portuguese}

\author{%
Thales Sales Almeida$^{1,3}$ \and
Ramon Pires$^{1,3}$ \and
Hugo Abonizio$^{2,3}$ \and
Rodrigo Nogueira$^{2,3}$ \and
Hélio Pedrini$^{1}$\and\\[0.5em]\small
$^{1}$ Institute of Computing, University of Campinas (UNICAMP)\\
$^{2}$ School of Electrical and Computer Engineering, University of Campinas (UNICAMP)\\
$^{3}$ Maritaca AI }

\begin{document}

\maketitle

\begin{abstract}
Large Language Models (LLMs) exhibit significant variations in performance across linguistic and cultural contexts, underscoring the need for systematic evaluation in diverse languages. In this work, we present the most extensive evaluation of LLMs for the Portuguese language to date. Leveraging our newly introduced PoETa~v2 benchmark -- a comprehensive suite of over 40 tasks in Portuguese -- we assess more than 20 models covering a broad spectrum of training scales and computational resources. Our study reveals how computational investment and language-specific adaptation impact performance in Portuguese, while also analyzing performance gaps in comparison to equivalent tasks in English. Through this benchmark and analysis, PoETa~v2 lays the groundwork for future research on Portuguese language modeling and evaluation. The benchmark is available at \url{https://github.com/PoETaV2/PoETaV2}.
\end{abstract}

\section{Introduction}

The rapid advancement of LLMs has fundamentally transformed natural language processing, enabling substantial progress across a wide range of applications~\cite{llm_applications1,llm_application3,llm_aplication4_law,llm_application_2_med}. However, as LLMs are increasingly deployed worldwide, it has become evident that their performance varies significantly across languages and cultural contexts~\cite{tiebe,worldbench,blend}. Linguistic diversity and uneven representation in training data mean that many languages, including Portuguese, remain underrepresented, often resulting in lower model performance.

This uneven landscape has created critical gaps in our understanding of LLM capabilities in lower-resource languages. Comprehensive, region-specific evaluation is therefore essential, both to accurately measure model performance and to guide the development of LLMs that effectively serve diverse communities.

In this work, we address this gap for Portuguese by introducing PoETa~v2 (Portuguese Evaluation Tasks), a benchmark for systematic and wide-ranging evaluation of LLMs in Portuguese. PoETa~v2 comprises a total of 44 tasks, 12 of which are natively designed in Portuguese and capture regional knowledge, culture, and linguistic features, such as tasks based on local exams, proverbs, and colloquial expressions. Additionally, we include 32 translated tasks adapted from established English benchmarks, ensuring comprehensive coverage of linguistic skills and domains.

Using PoETa~v2, we conduct the largest evaluation of LLMs in Portuguese to date, assessing the performance of over 20 prominent open-source and proprietary models spanning a wide range of sizes and training budgets. Our large-scale analysis provides a detailed characterization of LLM performance in Portuguese, highlighting the influence of model scale, computational investment, and language-specific pretraining.

\section{Related Work}

In this section, we review prior work in three key areas: (i) regional performance variation in LLMs, (ii) development of Portuguese-specialized models, and (iii) existing evaluation tasks for Portuguese.

\subsection{Regional Variation in LLM Performance}

Recent studies have shown that LLM performance can differ considerably depending on cultural and regional context. The Blend~\cite{blend} benchmark measures language models' understanding of daily activities in different countries, demonstrating that models are more attuned to regions with stronger digital representation. Similarly, WorldBench~\cite{worldbench} leverages socioeconomic data from the World Bank to evaluate how well models recall information about different countries, revealing higher error rates in underrepresented regions, such as Africa and the Middle East.

TiEBe~\cite{tiebe} evaluates factual recall of notable events across countries by generating QA pairs from Wikipedia retrospective pages. Results indicate that LLMs perform better for events in economically stronger countries, with a high correlation observed between a country's GDP and model performance. These findings reveal systematic biases in LLM capabilities and reinforce the importance of regionally aware evaluation initiatives, such as PoETa~v2 for Portuguese-speaking countries.

\subsection{Specialized Portuguese LLMs}

Although research on Large Language Models (LLMs) has primarily focused on high-resource languages~\cite{longpre2024bridging}, there have been several dedicated efforts to develop models specialized for Portuguese. These models vary in architecture, training paradigms, and scale, reflecting both the linguistic diversity of Portuguese and the computational resources available. We summarize the main Portuguese LLM initiatives as follows:

\textbf{Sabiá series}~\cite{pires2023sabia, sabia2, sabia3}: The Sabiá models represent some of the earliest attempts to adapt LLMs to Portuguese. The first generation includes Sabiá-7B and Sabiá-65B (both Llama-based) and Sabiá-J (GPT-J-based), trained on approximately 7.8 billion tokens from a carefully filtered subset of the ClueWeb 2022 dataset. These models were designed to capture a broad spectrum of Portuguese language patterns, including both European and Brazilian variants. Subsequent generations, Sabiá-2 and Sabiá-3, further improved model performance, but details of their methodology and training are limited. Access to these newer iterations is primarily via commercial APIs.

\textbf{Cabrita}~\cite{larcher2023cabrita}: Cabrita is based on a continued pretraining of the OpenLlama~3B model. The project adapted the tokenizer to better handle Portuguese-specific orthography and expanded the embedding layers to accommodate a larger vocabulary. Training involved 7 billion tokens extracted from the Portuguese subset of mC4.

\textbf{Glória}~\cite{lopes2024gloria}: Glória models, built on the GPTNeo architecture, include versions with 1.3B and 2.7B parameters, trained specifically for European Portuguese. Their dataset comprises 35.5 billion tokens, emphasizing formal and literary texts as well as contemporary usage. These models remain restricted to research access, but provide an important reference for evaluating Portuguese LLMs in the European variant.

\textbf{TeenyTinyLlama}~\cite{correa2024teenytinyllama}: TeenyTinyLlama is a family of lightweight models with 160M and 460M parameters, trained from scratch on 6.2 billion Portuguese tokens. Designed for resource-constrained environments, these models are suitable for deployment in scenarios where computational efficiency is critical, while still providing strong baseline performance across common NLP tasks.

\textbf{Tucano}~\cite{correa2024tucano}: The Tucano family includes models trained from scratch on up to 500 billion Portuguese tokens from the GigaVerbo dataset, with sizes up to 2.4B parameters. Tucano represents the largest monolingual Portuguese LLMs trained entirely from scratch to date. These models aim to capture extensive linguistic patterns across different Portuguese varieties and domains, from social media to formal literature.

\textbf{Curió}~\cite{curio}: Curió models are continually pretrained versions of existing LLMs, including Curió 1.1B (from TinyLlama) and Curió~7B (from LLama~2 7B). Both models were trained with up to 150 billion Portuguese tokens drawn from the ClassiCC-PT dataset, which aggregates diverse sources such as news articles, academic texts, and web crawls. Curió~7B is noteworthy for being the only Portuguese model at the 7B scale trained on over 100 billion Portuguese tokens, making it the largest Portuguese LLM in terms of both model size and training corpus.

Despite the increasing number of Portuguese-specialized LLMs, systematic comparison remains challenging. Each model is typically evaluated on a limited set of tasks or benchmarks, often with a limited number of examples. Consequently, it is difficult to assess the actual progress in Portuguese language modeling and to identify which models are best suited for specific applications. This limitation highlights the importance of large-scale, standardized evaluation frameworks such as PoETa~v2, which provide consistent and culturally relevant benchmarks across Portuguese LLMs.

\subsection{Existing Evaluation Tasks in Portuguese}

Evaluation resources for Portuguese LLMs have grown but remain fragmented. Approaches fall into two broad categories:

First, structured exams such as the Brazilian National University Entrance Exam (ENEM)~\cite{enem_challenge}, other university entrance exams~\cite{almeida2023bluex}, and the national bar exam~\cite{delfino2017passing, pires2025automatic} provide standardized, human-relevant assessments of reasoning and language understanding, but focus on specific academic or legal contexts.

Second, classification tasks derived from social media and informal text—such as TweetSentBR~\cite{tweetsentbr} for sentiment analysis, HateBR~\cite{vargas2021hatebr}, and PT Hate Speech~\cite{pt_hate_speech}—assess model performance on user-generated text and social signals.

Some exceptions, such as Faquad~\cite{sayama2019faquad} (extractive QA) and BRoverbs~\cite{broverbs} (multiple-choice on Brazilian proverbs), expand the task diversity.

Nevertheless, substantial gaps remain. There is a lack of comprehensive benchmarks for ethical reasoning, advanced problem-solving, code understanding, and cultural nuance in Portuguese. Existing datasets often do not cover commonsense reasoning, multilingual code generation, or ethical decision-making at the level of English benchmarks. Addressing these gaps is crucial for building robust, fair, and culturally aware Portuguese LLMs.

\section{Methodology}

To systematically evaluate the capabilities of LLMs in Portuguese, we designed a methodology that combines careful task selection, comprehensive model coverage, and standardized evaluation metrics. Our approach emphasizes both cultural specificity and general linguistic ability, enabling comparisons across models of different architectures, sizes, and training strategies. In the following subsections, we describe how PoETa~v2 was constructed, how tasks were categorized, the models included in our study, and the metrics used to assess their performance.

\subsection{Task Collection and Benchmark Construction}

PoETa~v2 was designed to provide a broad assessment of LLM performance in Portuguese, balancing regional specificity with general language capabilities. When curating the benchmark tasks, we prioritized those that could be solved using only pretraining knowledge, without requiring supervised fine-tuning. Consequently, we excluded open-ended chat tasks, code generation benchmarks, complex reasoning challenges, and agentic evaluations. The guiding principle was to construct a benchmark capable of tracking progress and guiding the development of Portuguese LLMs during the early stages of pretraining.

The tasks in PoETa~v2 are divided into two main classes:

\begin{itemize}

\item \textbf{Native Portuguese Tasks (12 tasks):} These tasks were developed natively in Portuguese and grounded in regional and cultural contexts. Examples include questions from Brazilian university entrance exams, tasks involving local proverbs and idiomatic expressions, and region-specific knowledge. Incorporating these tasks allows PoETa~v2 to evaluate how well models capture cultural and regional knowledge. Despite their importance, the limited number of native benchmarks means some fundamental aspects remain uncovered.

\item \textbf{Translated Tasks (32 tasks):} To broaden the evaluation, PoETa~v2 also includes tasks originally created in English and translated into Portuguese. These cover skills and domains not fully represented in the native tasks, such as general reasoning, reading comprehension, and commonsense understanding. While translated tasks enhance coverage, we note their limitations and emphasize the need for continued community efforts to increase the proportion of natively developed Portuguese tasks.

\end{itemize}

\subsection{Task Distribution}

Each task in PoETa~v2 is manually annotated with a primary task type to facilitate systematic evaluation. The main task types are defined as follows:

\begin{itemize}

\item \textbf{Regression:} Predicts a continuous numerical value rather than assigning the input to a discrete category. For example, given two sentences, the model assigns a similarity score between 1 and 5.

\item \textbf{Multiple-choice:} The model selects the correct answer from a set of predefined options. This format is commonly used in standardized exams and numerous benchmarks. We adopt the notation ``Multiple-choice (X)'' to indicate that a task offers X alternative options.

\item \textbf{Binary QA (Question Answering):} Provides a binary response to a question, such as ``yes'' or ``no.''

\item \textbf{Sentence entailment:} Determines whether a hypothesis sentence logically follows from a given premise.

\item \textbf{Extractive QA (Question Answering):} Identifies and extracts the correct answer span directly from a provided context passage in response to a question.

\item \textbf{Classification:} Assigns an input to one of several predefined, discrete classes. For example, classifying a news article into genres such as sports, technology, or politics.

\item \textbf{Sentiment Analysis:} A specialized form of classification that determines the sentiment expressed in a text. Examples include evaluating the sentiment of product reviews or social media posts.

\end{itemize}

In addition to the primary task type, each task in PoETa~v2 is annotated with one or more subcategories, providing finer-grained insight into the specific skills and capabilities the task is designed to evaluate. Examples of such subcategories include ``math'', for tasks that primarily assess mathematical reasoning, and ``Brazil'', for tasks that focus on knowledge or linguistic phenomena specific to Brazil.

\begin{figure}[!htb]
\centering
\subfloat[Distribution of task types in PoETa~v2]{\includegraphics[width=8.5cm]{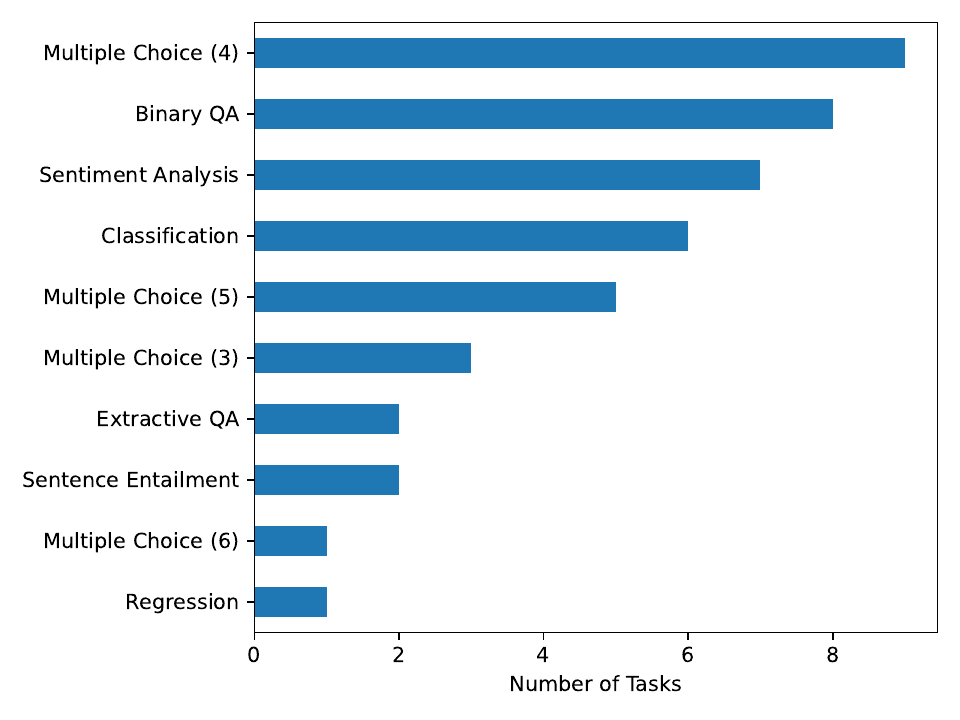} \label{fig:subfig1}} \\
\subfloat[Distribution of task categories in PoETa~v2]{\includegraphics[width=8.5cm]{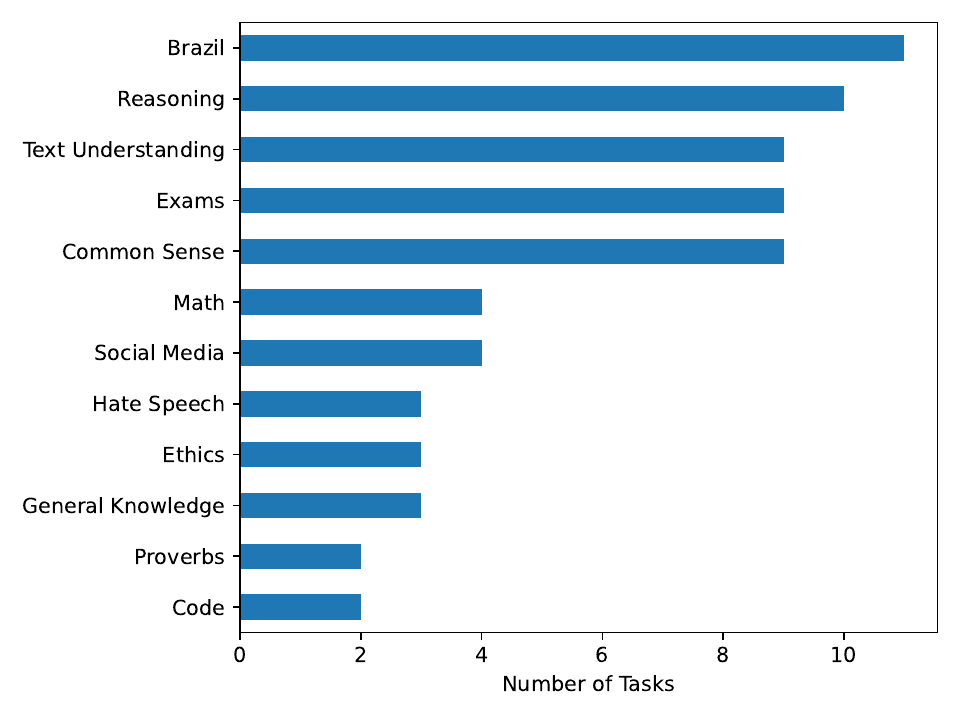} \label{fig:subfig2}}
\caption{Overview of the distribution of task types and categories in PoETa~v2. Each task is assigned a primary type and may belong to multiple subcategories.}
\label{fig:poetav2_type_and_subcategory_dist}
\end{figure}

Figure~\ref{fig:poetav2_type_and_subcategory_dist} illustrates the distribution of both primary task types and subcategories across the benchmark. While each task is assigned a single primary type, multiple subcategories may apply. Detailed information on the categorization of every task can be found in Appendix~\ref{sec:appendix_task_categorization}.

\subsection{Model Selection}

We selected over 20 LLMs for evaluation, including both open-source and commercial models, covering a wide range of model sizes and training budgets. This enables analysis of how computational investment and training strategies affect Portuguese language capabilities. We also include models with additional pretraining on Portuguese-specific data to assess the impact of targeted adaptation.

\subsection*{General Open-Source Models}

We evaluated several widely-used open-source models to establish robust baselines:

\begin{itemize}

\item \textbf{Llama (Versions 1, 2, and 3)~\cite{llama1,llama2,llama3}:} Three generations of the Llama family were included, as they represent some of the most popular open-source LLMs. Including multiple versions allows us to track progress and improvements across successive generations, specifically with respect to Portuguese language performance.

\item \textbf{Qwen (Versions 1, 2, 2.5, and 3)~\cite{qwen1,qwen25,team2024qwen2,yang2025qwen3}:} The Qwen family has consistently demonstrated strong multilingual capabilities, making these models suitable candidates for evaluating cross-linguistic performance in Portuguese.

\item \textbf{Falcon 3\cite{Falcon3}:} We focus on the latest Falcon models, which offer explicit support for Portuguese. Earlier Falcon versions (1 and 2) were excluded because their parameter count exceeded our 14B threshold, which was set to maintain comparability across models.

\end{itemize}

\subsection*{Portuguese-Specialized Models}
To evaluate the impact of targeted adaptation, we include several models specifically designed for Portuguese:

\begin{itemize}

\item \textbf{Sabiá 7B~\cite{pires2023sabia}:} The first Llama-based model further pretrained on Portuguese (10B tokens). Sabiá 7B represents an early effort to specialize general-purpose open models for Portuguese, providing a strong baseline for language-specific adaptation.

\item \textbf{Curió (1.1B and 7B)~\cite{curio}:} The Curió models undergo further pretraining with a focus on Portuguese. Curió 1.1B continues pretraining from the TinyLlama 1T checkpoint using 150B Portuguese tokens from the ClassiCC-PT dataset, while Curió~7B is adapted from Llama~2 7B and trained on 100B Portuguese tokens from the same dataset. These models allow us to examine how additional pretraining on Portuguese data affects performance across scales.

\end{itemize}

The Tucano family, trained from scratch on up to 500B Portuguese tokens~\cite{correa2024tucano}, represents one of the few fully monolingual LLM efforts in Portuguese. These models are limited to 2.4B parameters and were trained on fewer than 1T tokens. As a result, their behavior differs from other evaluated models, particularly in few-shot prompting scenarios, likely due to undertraining or variations in data distribution. For these reasons, Tucano results are presented separately in Appendix~\ref{sec:tucano_discussion}.

\subsection{Evaluation Metrics}

To evaluate model performance across the diverse tasks in PoETa~v2, we employ a set of metrics that capture both effectiveness, measured by NPM, and efficiency, measured by training compute. These metrics allow us to compare models of varying sizes and training strategies, while accounting for differences in task formats and scoring scales. In the following subsections, we describe the two main metrics used in our analysis.

\subsubsection{Computational Cost}

To estimate the computational resources required to train each open-source model in our benchmark, we adopt a slightly modified version of the non-embedding FLOPs per token metric proposed by DeepSeek~\cite{bi2024deepseek}. Specifically, the number of multiply–accumulate operations per token is defined as:
\begin{equation*}
M = 72 n_{\text{layer}} d_{\text{model}}^{2} + 12 n_{\text{layer}} d_{\text{model}}
\end{equation*}
and the overall training cost as:
\begin{equation*}
C = M \cdot D.
\end{equation*}
\noindent where $C$ denotes the total computational cost, $D$ is the number of training tokens (in trillions), $n_{\text{layer}}$ is the number of transformer layers, and $d_{\text{model}}$ is the model width.

Our formulation differs from the original DeepSeek metric by omitting the sequence length. This simplification accounts for recent models, such as Llama~3 and Qwen~3, which are pretrained in multiple stages with varying sequence lengths.

\subsubsection{Normalized Preferred Metric}

PoETa~v2 includes a diverse set of tasks, each with its own scoring range and evaluation criteria. To allow aggregation and comparison across tasks, we adopt the Normalized Preferred Metric (NPM), originally introduced in PoETa v1~\cite{pires2023sabia}:
\begin{equation*}
\text{NPM} = \frac{\text{[preferred metric]} - \text{[random score]}}{\text{[max score]} - \text{[random score]}}
\end{equation*}

NPM scales each task score such that a random model achieves an NPM of 0 and a perfect model achieves an NPM of 1. For example, in a four-option multiple-choice task, random guessing yields 25\% accuracy, while for an open-ended question-answering task, random performance is effectively 0\%. Simply averaging raw scores across tasks would assign a random model an average value of 12.5\%, obscuring meaningful performance differences. In contrast, NPM ensures a consistent baseline of 0, enhancing interpretability and enabling fair aggregation across heterogeneous tasks. This metric can accommodate various preferred evaluation measures, including accuracy, F1-score, and exact match, ensuring comparability across different task types.

\section{Results}

Table~\ref{tab:main_models_results} presents the average NPM across all PoETa~v2 tasks for more than 20 evaluated LLMs. The table also reports model performance on the subset of tasks derived from BIG-Bench, both in the original English version (BigBench EN) and the translated Portuguese version included in PoETa~v2 (BIG-Bench PT). Additionally, Table~\ref{tab:main_models_results} provides key model characteristics, including model size (in billions of parameters) and the total number of pretraining tokens, as reported by the respective model authors.

\begin{table*}[!htb]
\centering
\caption{Improvements in terms of NPM across categories for Portuguese-specialized LLMs.}
\label{tab:delta_specialized_models}
\begin{tabular}{@{}lccccccc@{}}
\toprule
& \multicolumn{3}{c}{\begin{tabular}[c]{@{}c@{}}Categories with the \\ Lowest Gain\end{tabular}} & \multicolumn{3}{c}{\begin{tabular}[c]{@{}c@{}}Categories with the \\ Highest Gains\end{tabular}}  \\
\midrule
& \begin{tabular}[c]{@{}c@{}}General\\ Knowledge\end{tabular} & Code &
Math & Brazil & Reasoning &  
\begin{tabular}[c]{@{}c@{}}Common\\ Sense\end{tabular} \\ \midrule
TinyLlama 1T $\rightarrow$ Curió 1.1B & -8.37 & -6.84  & 5.76 & -1.78 & 15.86  & 7.17   \\
Llama 1 7B $\rightarrow$ Sabiá 7B     & 7.47  & 19.15  & 19.15 & 19.42 & 8.80 & 15.95 \\
Llama 2 7B $\rightarrow$ Curió 7B     & 2.29  & -2.82 & -2.56 & 10.94 & 4.22  & 8.16 \\
\bottomrule
\end{tabular}
\end{table*}

\begin{table*}[!htb]
\centering
\caption{Results for all evaluated models on PoETa~v2, showing the average NPM across the full benchmark, as well as performance on the BIG-Bench subset in both Portuguese and English. The table also reports each model's pretraining corpus size and number of parameters.}
\label{tab:main_models_results}
\begin{tabular}{lcccccc}
\toprule
\multicolumn{1}{c}{Model} &
  \begin{tabular}[c]{@{}c@{}}Pretraining \\ Size\end{tabular} &
  \begin{tabular}[c]{@{}c@{}}Model \\ Size\end{tabular} &
  \multicolumn{1}{c}{\begin{tabular}[c]{@{}c@{}}Computational\\ Cost\end{tabular}} &
  \multicolumn{1}{c}{PoETa~v2} &
  \begin{tabular}[c]{@{}c@{}}BIG-Bench \\ (PT)\end{tabular} &
  \begin{tabular}[c]{@{}c@{}}BIG-Bench \\ (EN)\end{tabular} \\ \hline
\multicolumn{7}{c}{\textbf{Curió}}                                                                                                                           \\ \hline
\multicolumn{1}{l}{Curió 1.1B}    & 1T + 100B &  1.1B    & \multicolumn{1}{c}{7.3}  & \multicolumn{1}{c}{14.7} & 12.9 & 18.2 \\
\multicolumn{1}{l}{Curió 7B}      & 2T + 100B & 7B   & \multicolumn{1}{c}{81.2} & \multicolumn{1}{c}{34.8} & 30.9 & 27.2 \\ \hline
\multicolumn{7}{c}{\textbf{Sabiá}}                                                                                                                           \\ \hline
\multicolumn{1}{l}{Sabiá}         & 1T + 10B  & 7B   & \multicolumn{1}{c}{39.0}     & \multicolumn{1}{c}{32.4} & 26.3                 & 28.3                 \\ \hline
\multicolumn{7}{c}{\textbf{Llama}}                                                                                                                           \\ \hline
\multicolumn{1}{l}{TinyLlama 1T}  & 1T        & 1.1B & \multicolumn{1}{c}{6.64}   & \multicolumn{1}{c}{10.2} & 5.6 & 15.2 \\
\multicolumn{1}{l}{Llama 1 7B}    & 1T        & 7B   & \multicolumn{1}{c}{38.6}     & \multicolumn{1}{c}{19.9} & 14.0                   & 27.3                 \\
\multicolumn{1}{l}{Llama 2 7B}    & 2T        & 7B   & \multicolumn{1}{c}{77.3}    & \multicolumn{1}{c}{29.5} & 29.3                 & 29.7                 \\
\multicolumn{1}{l}{Llama 2 13B}   & 2T        & 13B  & \multicolumn{1}{c}{151}    & \multicolumn{1}{c}{41.5} & 41.3                 & 43.0                   \\
\multicolumn{1}{l}{Llama 3.1 8B}  & 15T       & 8B   & \multicolumn{1}{c}{580}   & \multicolumn{1}{c}{53.5} & 50.2                 & 52.8                 \\ \hline
\multicolumn{7}{c}{\textbf{Falcon 3}}                                                                                                                          \\ \hline
\multicolumn{1}{l}{Falcon 3 1B}   & 14T       & 1B   & \multicolumn{1}{c}{76}    & \multicolumn{1}{c}{18.6} & 20.5                 & 31.0                 \\
\multicolumn{1}{l}{Falcon 3 3B}   & 14T       & 3B   & \multicolumn{1}{c}{209}    & \multicolumn{1}{c}{38.6} & 37.4                 & 45.1                 \\
\multicolumn{1}{l}{Falcon 3 7B}   & 14T       & 7B   & \multicolumn{1}{c}{266}    & \multicolumn{1}{c}{58.5} & 55.2                 & 55.9                 \\
\multicolumn{1}{l}{Falcon 3 10B}  & 14T       & 10B  & \multicolumn{1}{c}{380}   & \multicolumn{1}{c}{63.5} & 60.4                 & 62.8                 \\ \hline
\multicolumn{7}{c}{\textbf{Qwen~1 and~2}}                                                                                                                            \\ \hline
\multicolumn{1}{l}{Qwen~1 1.8B}   & 2.4T      & 1.8B & \multicolumn{1}{c}{17.4}  & \multicolumn{1}{c}{19.0} & 21.3                 & 30.4                 \\
\multicolumn{1}{l}{Qwen~1 7B}     & 2.4T      & 7B   & \multicolumn{1}{c}{93.8}  & \multicolumn{1}{c}{41.1} & 36.9                 & 43.8                 \\
\multicolumn{1}{l}{Qwen~2 1.5B}   & 7T        & 1.5B & \multicolumn{1}{c}{33.3}  & \multicolumn{1}{c}{38.4} & 35.1                 & 34.5                 \\
\multicolumn{1}{l}{Qwen~2 7B}     & 7T        & 7B   & \multicolumn{1}{c}{181.3}    & \multicolumn{1}{c}{58.8} & 51.8                 & 56.2                 \\ \hline
\multicolumn{7}{c}{\textbf{Qwen~2.5}}                                                                                                                            \\ \hline

\multicolumn{1}{l}{Qwen~2.5 1.5B} & 18T       & 1.5B & \multicolumn{1}{c}{85.5}    & \multicolumn{1}{c}{43.0} & 37.0                 & 39.7                 \\
\multicolumn{1}{l}{Qwen~2.5 3B}   & 18T       & 3B   & \multicolumn{1}{c}{195.7}    & \multicolumn{1}{c}{53.1} & 46.8                 & 50.5                 \\
\multicolumn{1}{l}{Qwen~2.5 7B}   & 18T       & 7B   & \multicolumn{1}{c}{466.2}   & \multicolumn{1}{c}{63.7} & 57.4                 & 58.0                 \\
\multicolumn{1}{l}{Qwen 2.5 14B}   & 18T       & 14B   & \multicolumn{1}{c}{1630.8}   & \multicolumn{1}{c}{71.0} & 63.5                 & 62.0                 \\

\hline
\multicolumn{7}{c}{\textbf{Qwen~3}}                                  \\ \hline
\multicolumn{1}{l}{Qwen~3 1.7B}   & 36T       & 1.7B & \multicolumn{1}{c}{304}  & \multicolumn{1}{c}{45.1} & 44.9 & 46.1 \\
\multicolumn{1}{l}{Qwen~3 4B}     & 36T       & 4B   & \multicolumn{1}{c}{611}   & \multicolumn{1}{c}{59.2} & 56.6 & 57.9 \\
\multicolumn{1}{l}{Qwen~3 8B}     & 36T       & 8B   & \multicolumn{1}{c}{1565}   & \multicolumn{1}{c}{64.7} & 57.7 & 63.2 \\
\multicolumn{1}{l}{Qwen~3 14B}    & 36T       & 14B  & \multicolumn{1}{c}{2717}   & \multicolumn{1}{c}{70.5} & 64.7 & 74.2 \\ \hline

\multicolumn{7}{c}{\textbf{Commercial Models}}                                  \\ \hline

\multicolumn{1}{l}{GPT 4.1}     & -       & -   & \multicolumn{1}{c}{-}   & \multicolumn{1}{c}{76.2} & 67.9 & 69.9 \\
\multicolumn{1}{l}{GPT-4o}     & -       & -   & \multicolumn{1}{c}{-}   & \multicolumn{1}{c}{75.2} & 67.7 & 72.2 \\
\multicolumn{1}{l}{Sabiá~3}    & -       & -  & \multicolumn{1}{c}{-}   & \multicolumn{1}{c}{72.2} & 62.8 & 59.4 \\
\bottomrule
\end{tabular}
\end{table*}

\subsection{Overall Performance on PoETa~v2}

\begin{figure*}[!htb]
\centering
\includegraphics[width=0.98\linewidth]{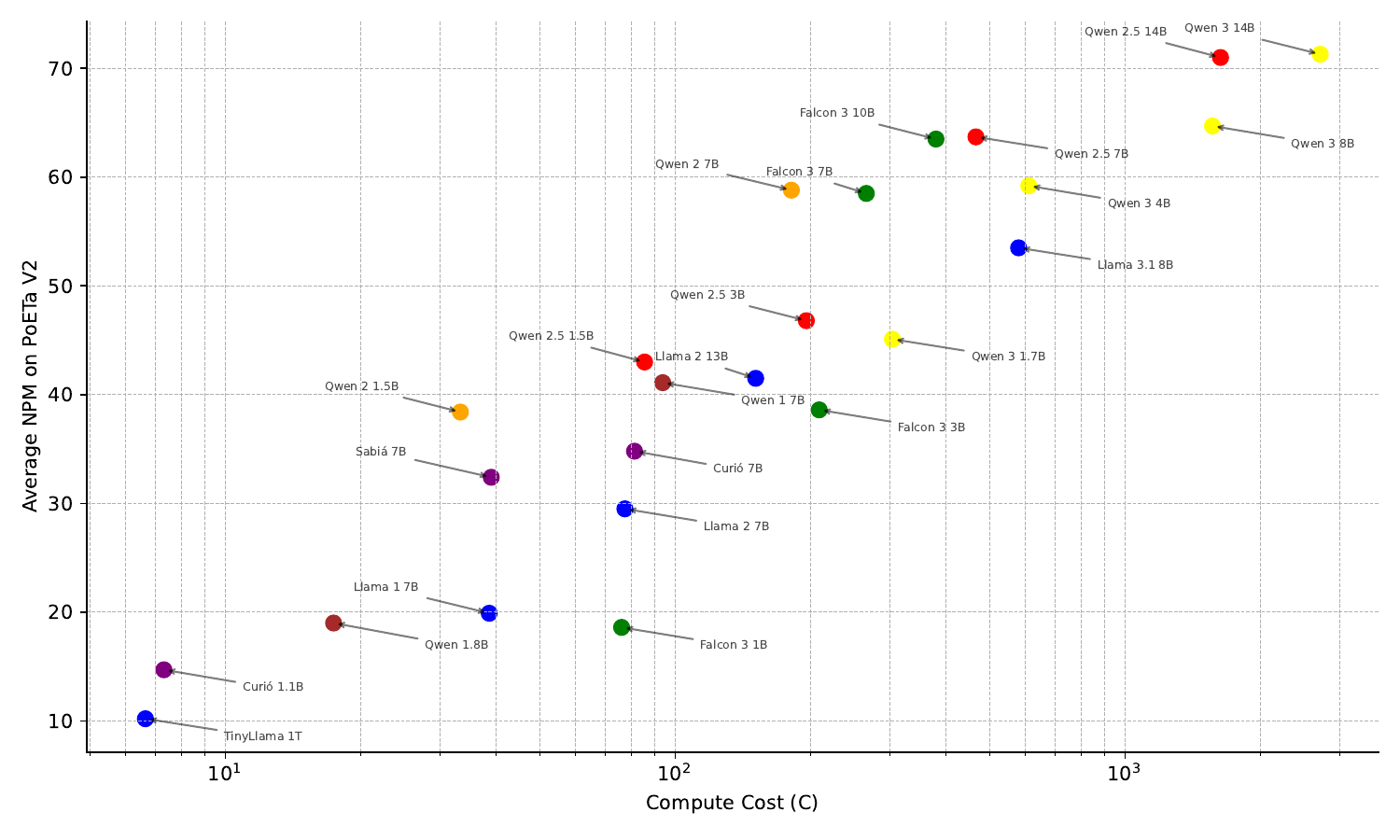}
\caption{Computational cost versus average NPM score for the evaluated models on PoETa~v2. Colors distinguish different model families.}
\label{fig:linear_scatter_plot}
\end{figure*}

Figure ~\ref{fig:linear_scatter_plot} shows a clear positive correlation between computational cost and performance on PoETa~v2. Models with larger numbers of parameters and more extensive pretraining corpora generally achieve higher average NPM scores, indicating that increasing both model size and training data contribute to improved performance in Portuguese.

Among open-source LLMs, the Qwen family stands out. Qwen~3 14B and Qwen~2.5 14B achieve the highest average NPM scores among the evaluated open-source models—70.5 and 71, respectively—outperforming all other models in the benchmark. Interestingly, Qwen~3 14B incurs roughly twice the computational cost of Qwen~2.5 14B, yet both reach comparable performance levels.

Although computational cost is generally a strong predictor of model quality on PoETa~v2, it does not capture the full picture. Notable outliers highlight the importance of architectural design and training data composition. For instance, Qwen~1 7B and Llama~2 7B exhibit similar computational costs, yet Qwen~1 7B surpasses Llama~2 7B by more than 12 points in average NPM. A similar trend is observed among higher-cost models: Llama~3.1 8B and Qwen~2.5 7B have comparable computational costs, but Qwen~2.5 7B outperforms by nearly 10 points.

These discrepancies indicate that factors beyond model scale -- such as architecture and the composition of pretraining data -- can have a significant impact. In particular, the Qwen models' superior performance in Portuguese may be attributable to a higher proportion of Portuguese-language data in their pretraining corpora compared to their Llama counterparts. Unfortunately, detailed information about the training data for these models is not publicly available, limiting the depth of our analysis. Nevertheless, our results strongly suggest that both computational investment and the linguistic characteristics of the pretraining data are crucial for optimizing LLM performance in Portuguese.

Among commercial systems, GPT-4.1 and GPT-4o establish the current performance ceiling on PoETa~v2, achieving average NPM scores above 75. While these models outperform all open-source alternatives, the margin is relatively small—approximately five points higher than Qwen~2.5 14B. Sabiá~3 also attains competitive results, with an average NPM of 72.2, slightly surpassing the top-performing Qwen models. The narrow gap between the leading open-source models and commercial systems suggests that PoETa~v2 may be nearing a saturation point, which aligns with the benchmark's design, as it emphasizes tasks that can be solved without extensive supervised fine-tuning.

\subsection{Language and Regional Specialization}

To evaluate the impact of continued pretraining on Portuguese data, we consider three models explicitly adapted in this manner: Curió 1.1B (initialized from TinyLlama), Curió 7B (initialized from Llama 2 7B), and Sabiá (initialized from Llama 1 7B). Both Curió models underwent further pretraining with approximately 100 billion Portuguese tokens, while Sabiá was pretrained with 10 billion Portuguese tokens.

All three models demonstrate notable improvements on PoETa~v2 relative to their respective base models. Specifically, Curió 1.1B gains 4.5 NPM points over TinyLlama, Curió 7B gains 5.3 points over Llama~2 7B, and Sabiá achieves a striking 12.5-point improvement over Llama~1 7B.

The magnitude of Sabiá's improvement is particularly remarkable given its smaller additional pretraining corpus—ten times smaller than that used for the Curió models. This effect can be explained by the performance of Llama~1 7B on the BIG-Bench subset of PoETa~v2: the base model achieves only half of its English score when evaluated in Portuguese, revealing a significant deficiency in handling Portuguese-language tasks. Sabiá, however, substantially mitigates this imbalance, reaching comparable performance in both languages after only 10 billion additional Portuguese tokens. This suggests that Sabiá's gains are primarily attributable to exposure to Portuguese content that was largely absent in the original Llama~1 7B, which explicitly filtered out non-English data during pretraining.

Interestingly, Llama~2 7B -- the base model for Curió~7B -- does not exhibit such a pronounced language imbalance, indicating that it already possessed stronger multilingual capabilities. Nevertheless, continued pretraining in Portuguese still yields meaningful improvements, underscoring the value of language-specific adaptation even for multilingual models.

\subsection{Performance Variance Between Portuguese and English}

Across the evaluated models, we observe an average performance difference of 3.5 points between BIG-Bench tasks in Portuguese and English, with most models achieving higher scores in English. Notable exceptions include Curió~7B and Qwen~2.5 14B, which perform slightly better in Portuguese.

A more detailed analysis reveals that model size influences the magnitude of this language gap. Smaller models (fewer than 5 billion parameters) exhibit an average difference of 5.1 points, while mid-sized models (7B–8B parameters) show a reduced gap of 4.3 points. For larger models with 10B parameters or more, the average difference further decreases to 3.8 points. This trend suggests that increased model scale not only enhances absolute performance but also narrows the gap between English and Portuguese, likely reflecting stronger multilingual capabilities.

Nonetheless, significant outliers exist. Qwen~3 8B and Qwen~3 14B display unusually large gaps of 5.5 and 9.5 points, respectively, performing exceptionally well on English BIG-Bench tasks but failing to achieve comparable results in Portuguese. This discrepancy may stem from differences in pretraining data composition or potential data contamination. However, due to the limited availability of detailed information about the Qwen~3 pretraining dataset, further analysis is not currently possible.

\subsection{Performance by Subcategories}

\begin{figure*}[!htb]
\centering
\subfloat[Brazil]{\includegraphics[width=0.31\linewidth]{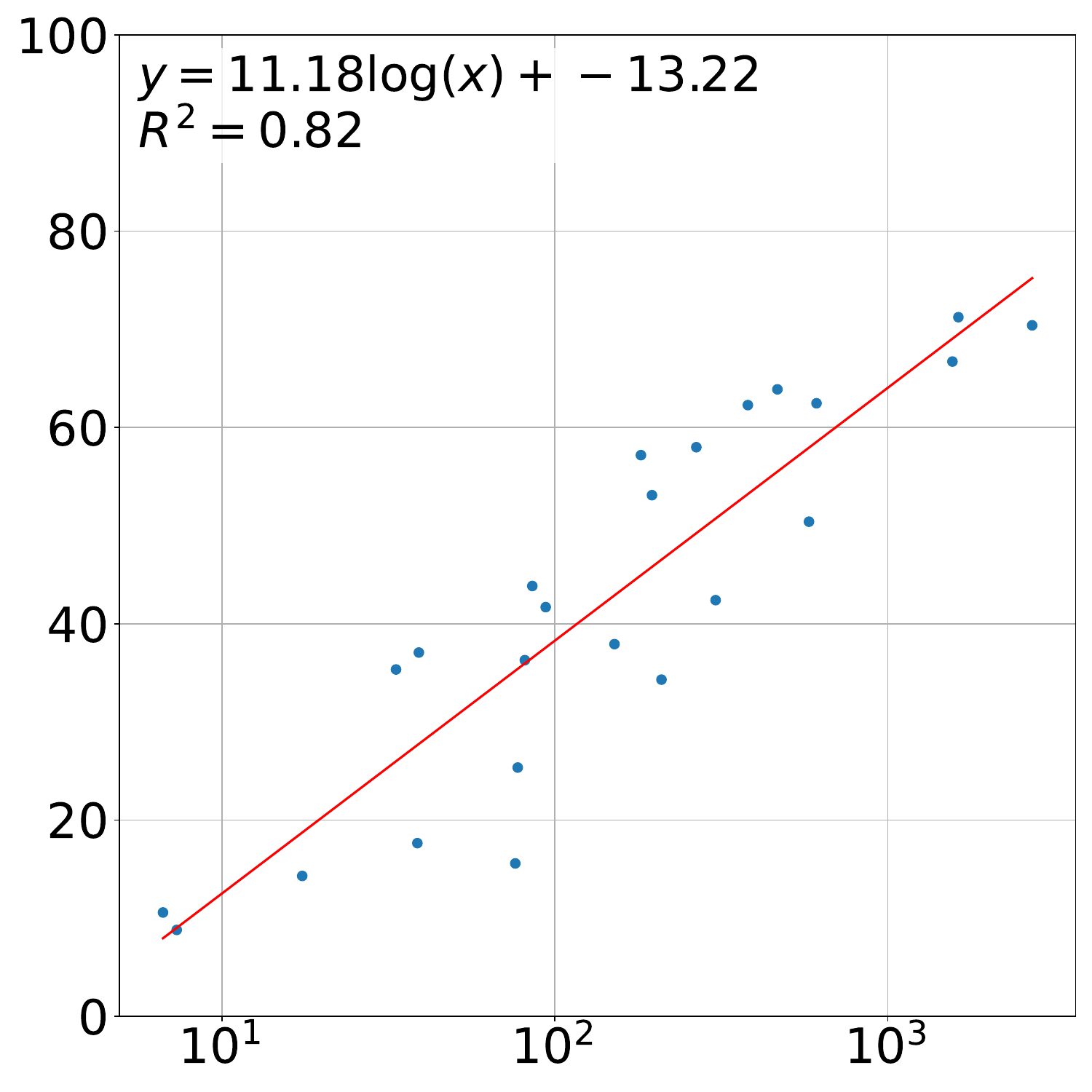}} \hspace*{0.5cm}
\subfloat[Text Understanding]{\includegraphics[width=0.31\linewidth]{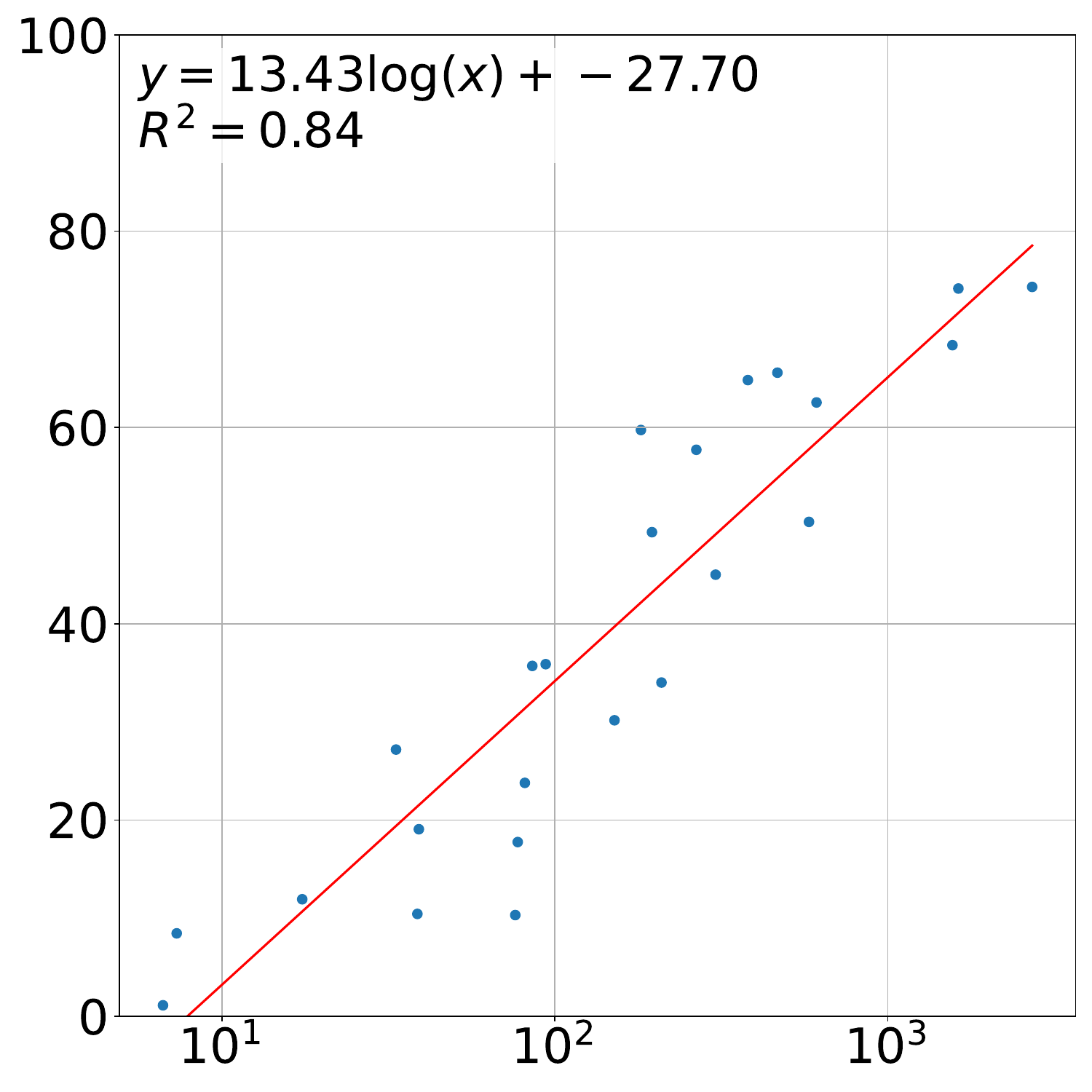}} \hspace*{0.5cm}
\subfloat[Exams]{\includegraphics[width=0.31\linewidth]{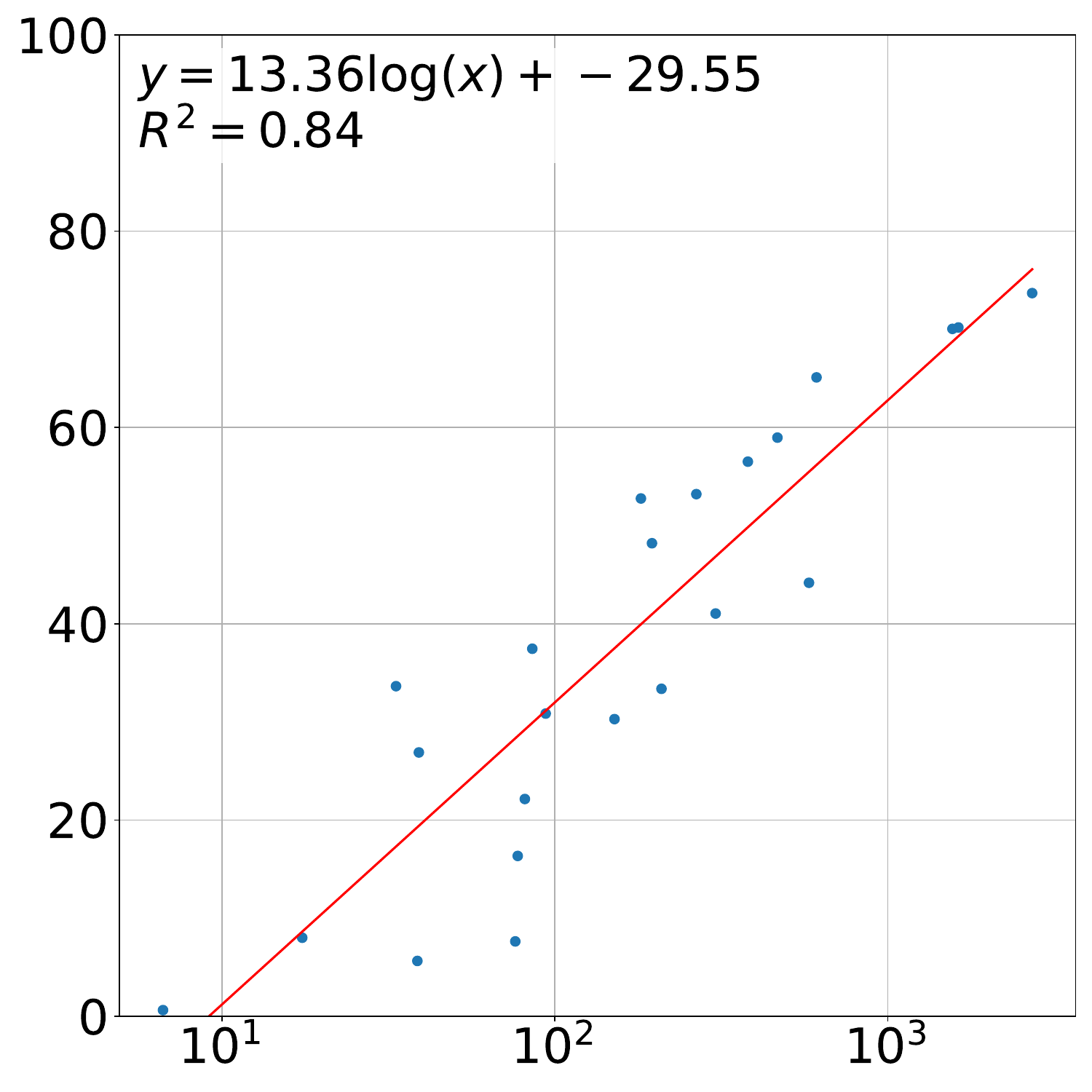}} \\   
\subfloat[Reasoning]{\includegraphics[width=0.31\linewidth]{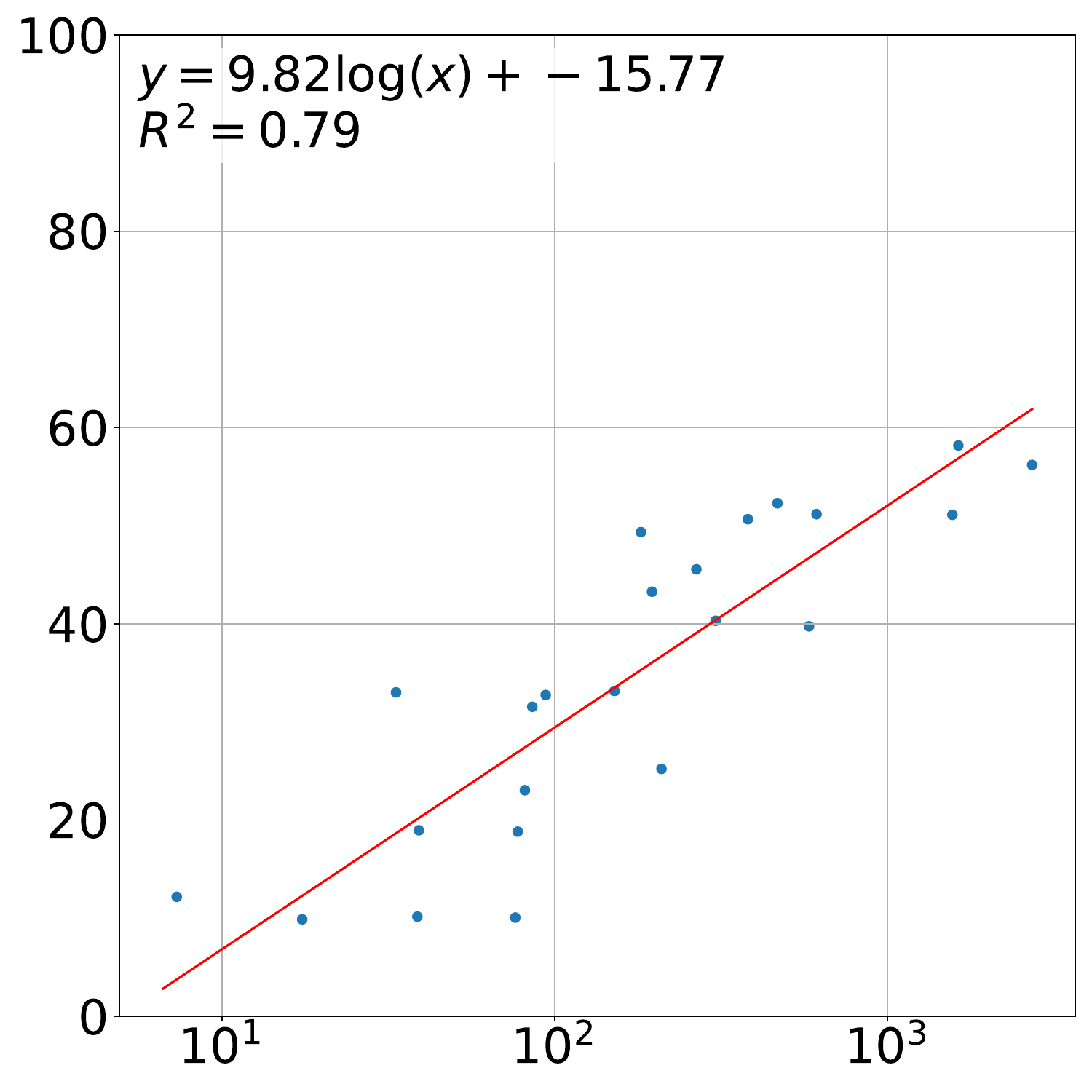}} \hspace*{0.5cm}
\subfloat[Common Sense]{\includegraphics[width=0.31\linewidth]{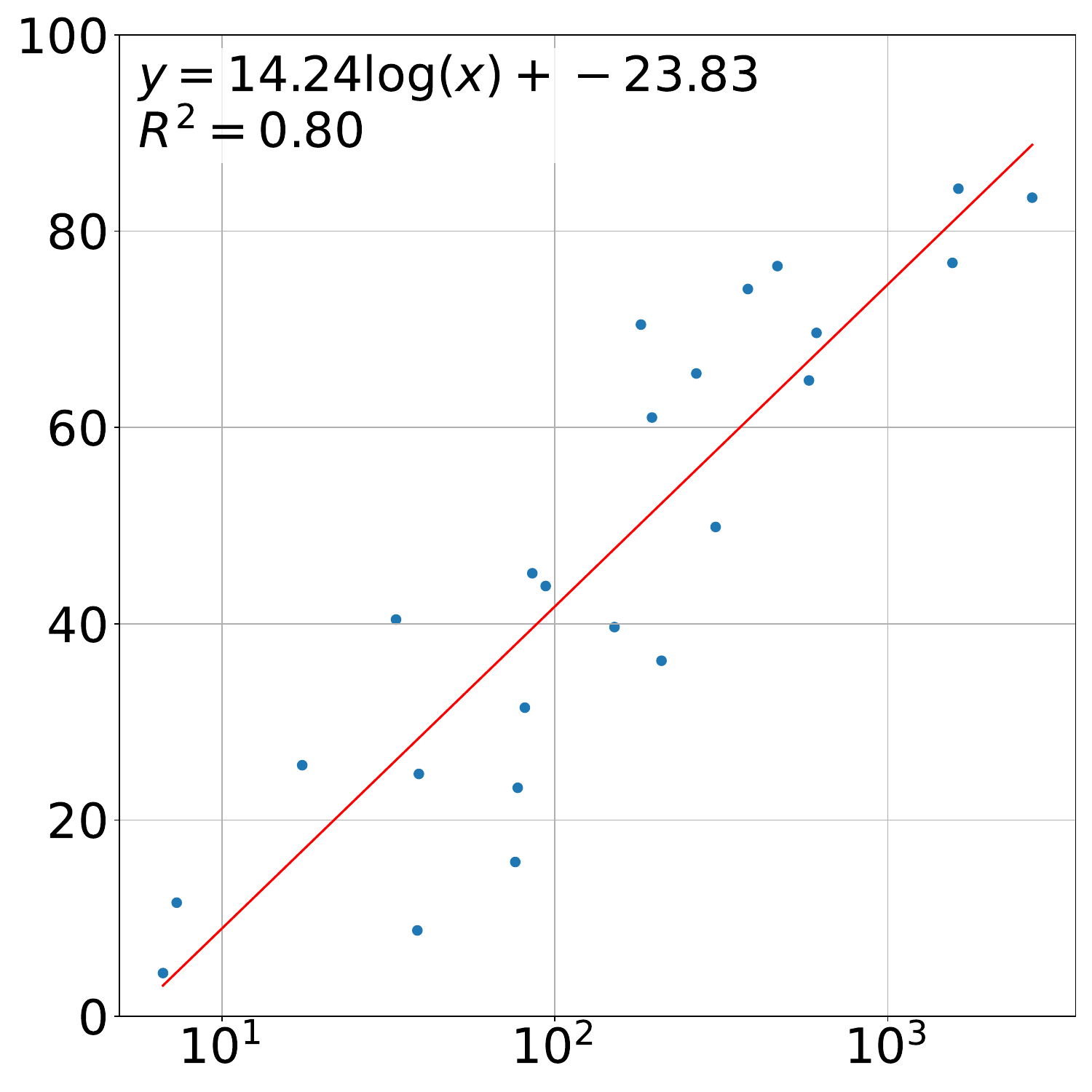}} \hspace*{0.5cm}
\subfloat[Math]{\includegraphics[width=0.31\linewidth]{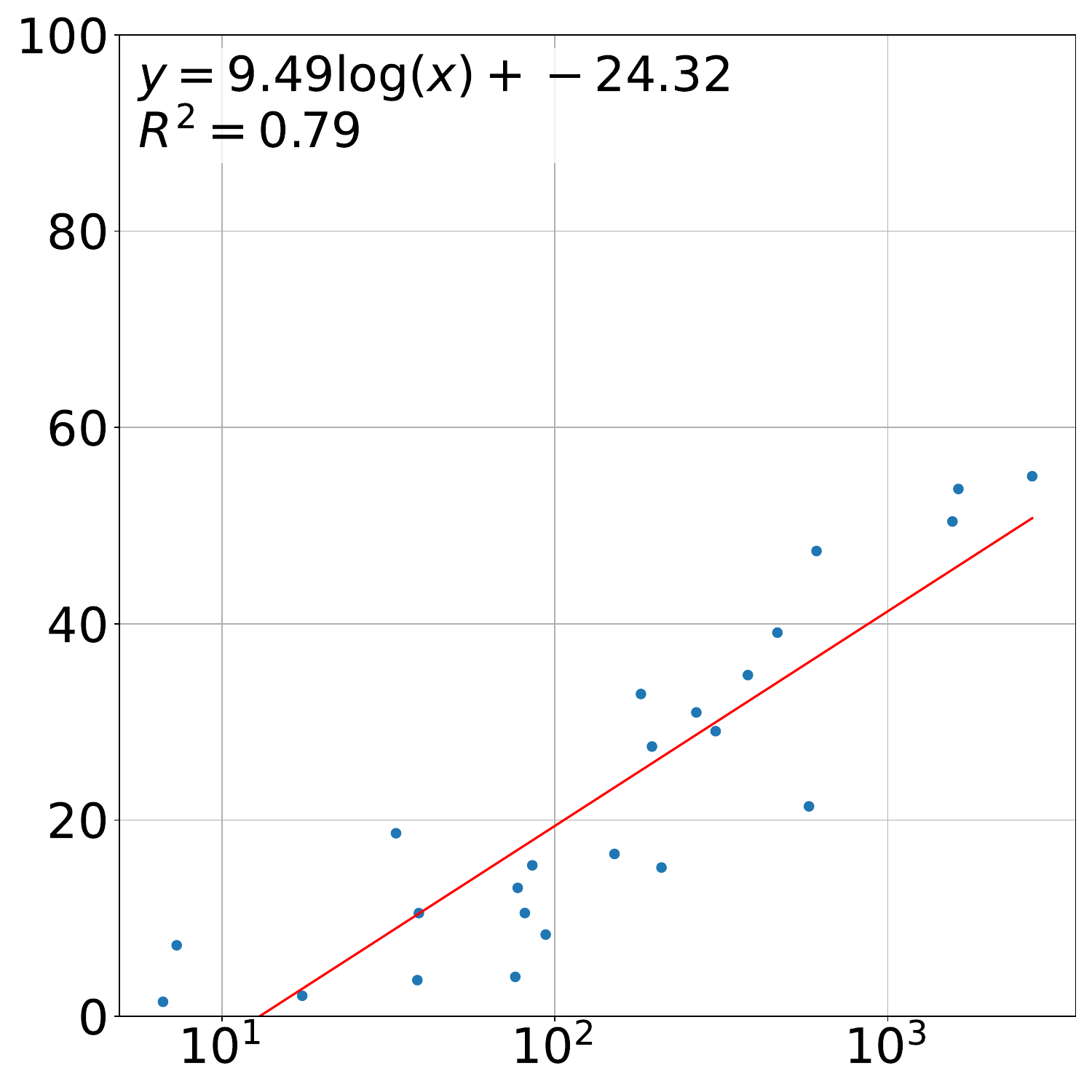}} \\
\subfloat[Social Media]{\includegraphics[width=0.31\linewidth]{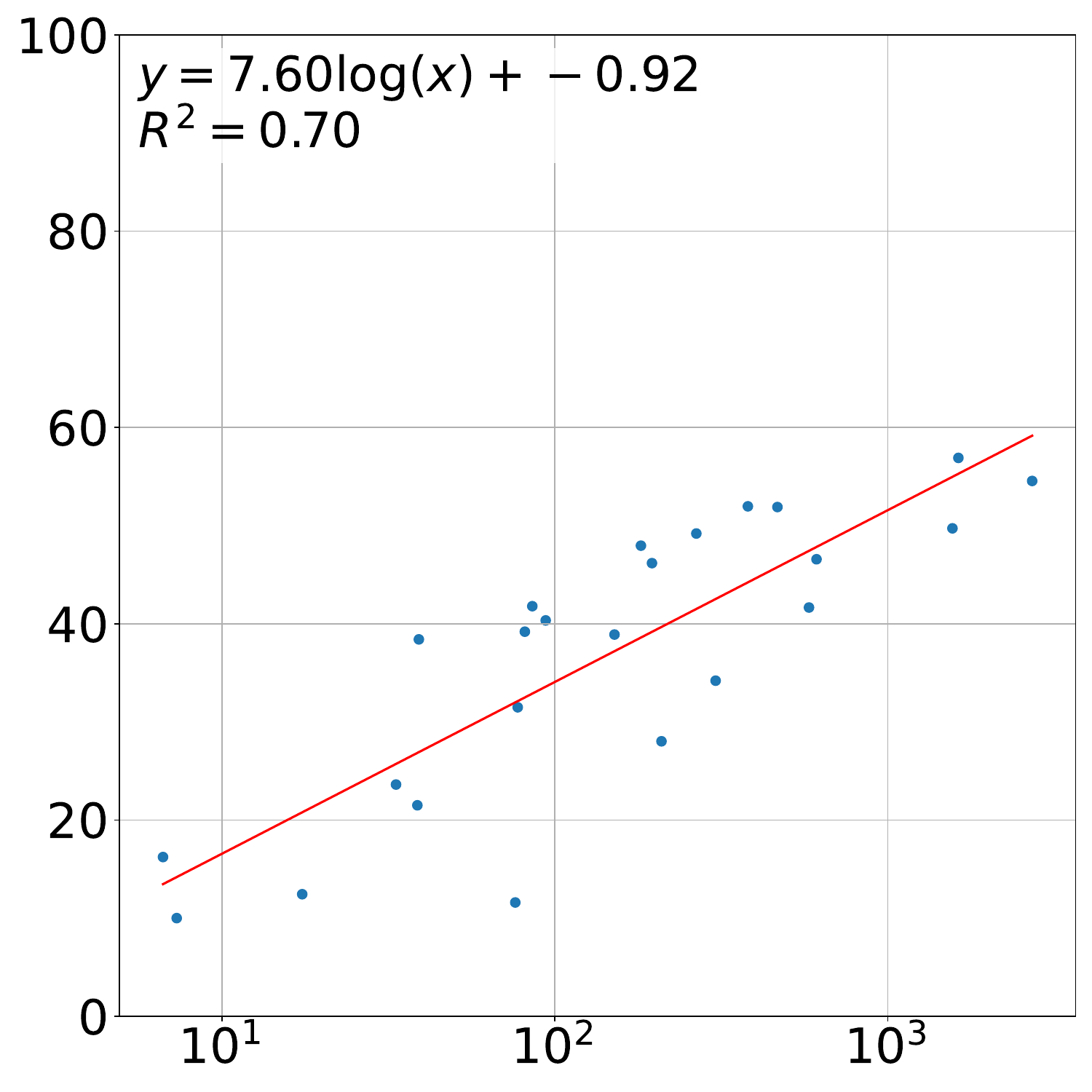}} \hspace*{0.5cm}
\subfloat[General Knowledge]{\includegraphics[width=0.31\linewidth]{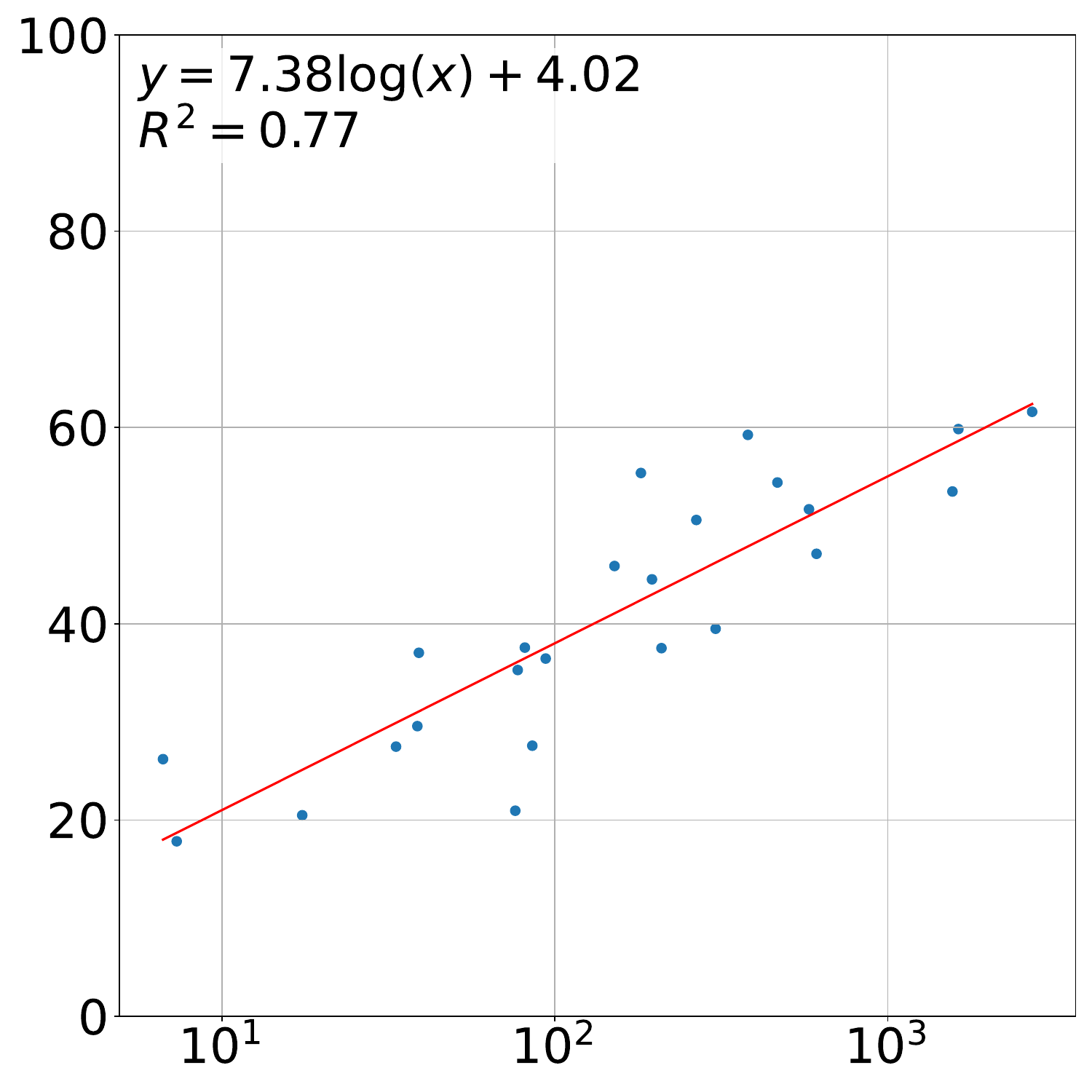}} \hspace*{0.5cm}
\subfloat[Ethics]{\includegraphics[width=0.31\linewidth]{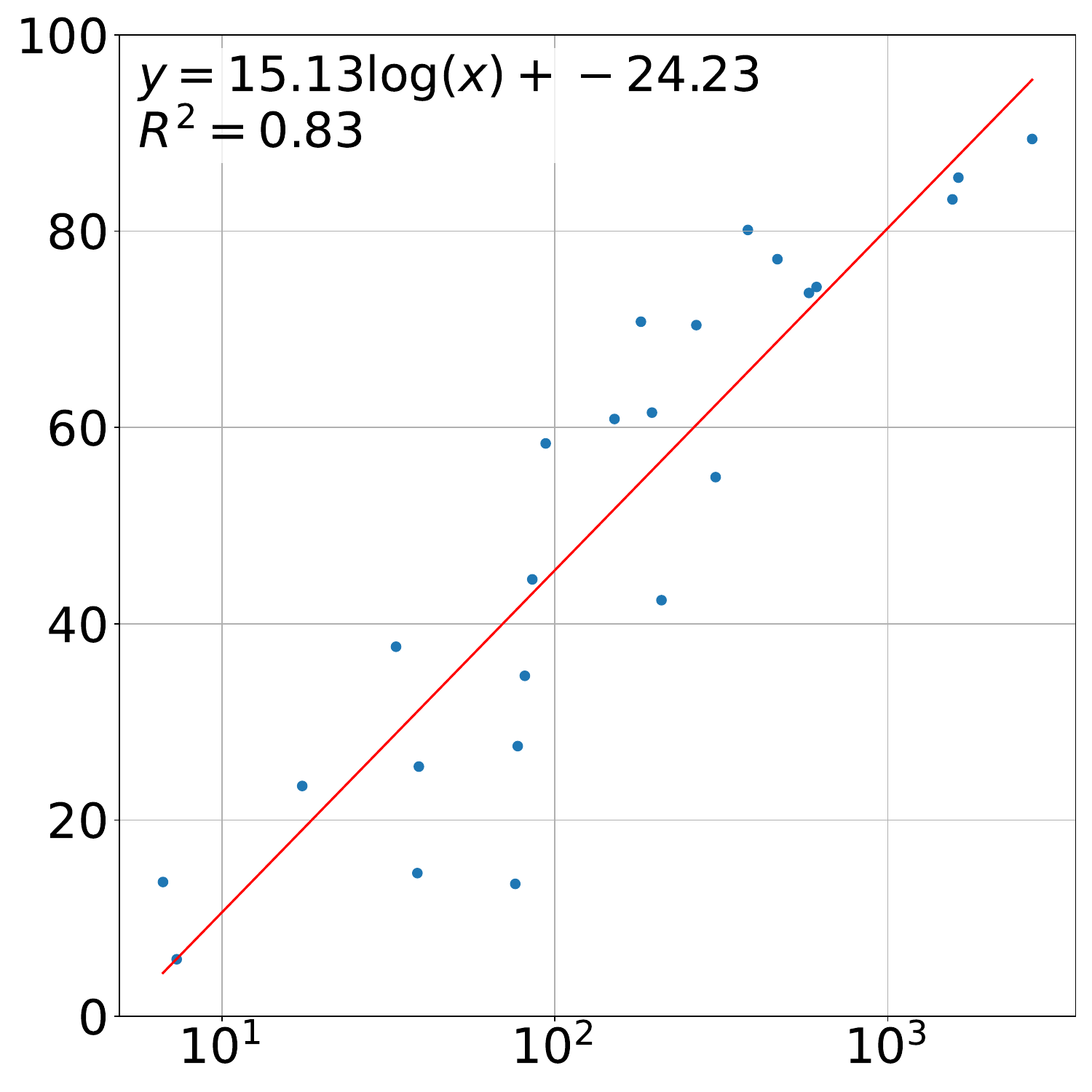}}
\caption{Average NPM across task subcategories plotted against computational cost.}
\label{fig:3x3grid_subcategories}
\end{figure*}

Figure~\ref{fig:3x3grid_subcategories} presents the relationship between computational cost and performance across the nine most recurrent subcategories in PoETa~v2.
Across all categories, we observe a positive correlation between computational cost and normalized performance metric (NPM), with $R^2$ values ranging from 0.70 (Social Media) to 0.84 (Text Understanding and Exams). This indicates that larger and more computationally expensive models generally achieve better results in every evaluated skill area, although the strength of this trend varies by category.

Tasks in \textbf{Text Understanding}, \textbf{Exams}, \textbf{Ethics}, \textbf{CommonSense}, and \textit{Brazil} display some of the strongest correlations ($R^2 \geq 0.82$ ), suggesting that scaling model size and pretraining data tend to yield consistent gains. These categories likely benefit from general language modeling improvements and broad world knowledge.

Meanwhile, \textbf{Reasoning}, and \textbf{Math} also show substantial positive trends ($R^2 = 0.84$). However, the slopes in these tasks are lower, suggesting more limited returns from increased compute.

The weakest correlations appear in \textbf{Social Media} ($R^2=0.70$) and \textbf{General Knowledge} ($R^2=0.77$), both of which also present shallower scaling trends ($\text{slope} \leq 8$). In Social Media, the high linguistic variability, prevalence of informal registers, and domain-specific vocabulary likely limit the gains achievable through scaling alone. For General Knowledge, performance may depend more on the explicit presence of relevant facts in the pretraining corpus than on overall model size.

Overall, while computational cost is a strong predictor of performance across all PoETa~v2 subcategories, the magnitude of scaling benefits is not uniform.

\section{Conclusions}

In this work, we introduced PoETa~v2, a comprehensive benchmark for evaluating large language models in Portuguese, including non-instruction-tuned ones, across a diverse set of tasks. Through the largest systematic evaluation of Portuguese LLMs conducted to date, we analyzed model performance in relation to computational investment and language-specific adaptation.

Our findings reveal a strong linear correlation between computational cost and performance in Portuguese, while also highlighting notable outliers, indicating that model architecture and the composition of pretraining data play critical roles. Further pretraining on Portuguese data consistently enhances model performance, particularly for base models with limited initial multilingual exposure. Nonetheless, persistent performance gaps between English and Portuguese remain, especially for smaller models.

Despite recent progress, our analysis underscores ongoing challenges: the need for more natively developed Portuguese tasks, greater transparency regarding pretraining data, and targeted efforts to mitigate linguistic and regional disparities. By providing an open, extensible benchmark along with comprehensive baseline results, PoETa~v2 aims to facilitate the development of more equitable and culturally aware language models for the Portuguese language.

\section{Future Work}

Our findings highlight several promising directions for future research:

\begin{itemize}

\item \textbf{Expansion of Native Tasks:} Despite progress, the benchmark remains heavily reliant on translated tasks. Increasing the proportion and diversity of natively developed Portuguese tasks -- especially in domains such as ethical reasoning, commonsense knowledge, code understanding, and advanced problem solving -- will be crucial for robust evaluation. Community-driven initiatives and collaboration with domain experts can accelerate the development of new culturally and linguistically rich datasets.

\item \textbf{Deeper Analysis of Pretraining Data:} Our results suggest that pretraining data composition plays a central role in model performance, since many models with similar pretraining compute cost achieve notably different performances in PoETa~V2, however, the lack of transparency regarding data sources limits deeper analysis. Future work should prioritize both improved reporting of training data statistics and efforts to systematically quantify the impact of different data distributions, particularly regarding Portuguese and other underrepresented languages.

\item \textbf{Robustness, Fairness, and Bias Analysis:} A thorough assessment of model fairness, bias, and robustness in the Portuguese context is still lacking. Future benchmarks and studies should explicitly measure and address potential biases and performance disparities across regional dialects, demographic groups, and sensitive domains.

\item \textbf{Benchmarking for fine-tuned models:} While PoETa~v2 focuses on general language understanding and reasoning, our study evaluates most models in their pre-training stage. Our tasks are reasonably simple and do not cover common use cases for fine-tuned models, such as advanced use cases like long text generation or agentic capabilities.

\end{itemize}

Ultimately, we hope PoETa~v2 will serve as a foundation for continued progress in Portuguese language model evaluation and facilitate movement towards more inclusive, regionally grounded NLP research.

\bibliographystyle{plainnat}
\bibliography{preprint}{}

\appendix

\section{Evaluation Tasks}

This section details the evaluation tasks used to assess model performance. Each task is presented as a separate benchmark, briefly explaining the underlying skill or knowledge it aims to measure. Unless otherwise noted, tasks are translated into Portuguese and evaluated using accuracy as the primary metric.

\subsection*{AG News}
AG News~\cite{agnews} is a news classification dataset. In this task, the model is presented with a news article, and needs to classify it in one of four categories:

\begin{itemize}
    \item World
    \item Sports
    \item Business
    \item Technology
\end{itemize}

\begin{examplebox}
\textbf{Question:} Given the following title and text, choose the most appropriate news category.
\textit{Title:} Microsoft to share code with governments
\textit{Text:} Government officials sought greater access to the program’s technical details to ensure that the software is compatible with other applications and does not conceal obvious security flaws. – The Washington Post

\textbf{Answer:} Technology.
\end{examplebox}

\subsection*{ASSIN 2 RTE}
ASSIN 2 Recognizing Textual Entailment (RTE) is a Portuguese-language benchmark for evaluating a model's ability to detect textual entailment between sentence pairs. The task is binary classification: entailment or non-entailment.

\begin{examplebox}
\textbf{Question:} Given the premise "The goal is being guarded by a hockey player wearing a yellow team jersey," is it true that "A hockey player in a yellow jersey is protecting the goal"?

\textbf{Answer:} Yes.

\end{examplebox}

\subsection*{ASSIN 2 STS}
ASSIN 2 Semantic Textual Similarity (STS) is a regression dataset that measures how similar two sentences are, based on a continuous score. It is commonly used for evaluating sentence embeddings and model understanding of semantic similarity in Portuguese.

\begin{examplebox}
\textbf{Question:} Classify the following pair of sentences as 1, 2, 3, 4, or 5.
\begin{enumerate}
\item Completely different sentences, about different topics.
\item Unrelated sentences, but about the same topic.
\item Somewhat related: may describe different facts but share some details.
\item Strongly related sentences, but differing in some details.
\item Sentences essentially mean the same thing.
\end{enumerate}

\textit{Sentence 1:} A man is outside throwing blades at a nearby target.
\textit{Sentence 2:} A man is inside the house throwing blades at a nearby target.

\textbf{Answer:} 3.

\textbf{Explanation:} Both sentences describe a man throwing blades at a target, but they differ in location (outside vs. inside the house). This makes them related, sharing key details, but not strongly equivalent.
\end{examplebox}

\subsection*{BLUEX}
BLUEX is a multiple-choice question answering benchmark for two Brazilian university entrance exams (USP and UNICAMP). It features questions at the high school level and contains more than 500 questions in total.

BLUEX also provides questions containing images; however, in this work, we use only the questions without images for our evaluations.

\begin{examplebox}
\textbf{Question:} Based on the text below and your knowledge, choose the correct alternative.

\textbf{Statement:}
The word Anthropocene appears today in the title of hundreds of books and scientific articles, in thousands of citations, and its use continues to grow in the media. Referring to the era when human actions began to cause biophysical changes on a planetary scale, the term was created in the 1980s and popularized in the 2000s. Groups of specialists found that these changes affected the Earth System, which had been in relative balance since the beginning of the Holocene, 11,700 years ago. To mark the beginning of this new era, such groups symbolically chose the year 1784, the moment of the improvement and popularization of the steam engine. This context also corresponds to the beginning of the Industrial Revolution and the use of fossil fuels.

\textbf{Alternatives:}
A. From the 18th century, with the Enlightenment, the belief in human superiority over nature was widely questioned, which reduced the impacts of human actions on the Planet throughout the 20th and 21st centuries.
B. From the Modern Era, anthropocentrism guided the belief in human superiority over nature; this idea was consolidated in 1784, remaining in the sciences to the present with the name Anthropocene.
C. Climate change, alteration of vegetation cover, and large-scale biodiversity loss have been humanity’s marks on the Planet since the maritime expansions of the 16th century, being of little concern to science.
D. With the popularization of the American lifestyle, there was an acceleration of changes caused by human action on Planet Earth, changes that had been occurring since 1784, with the Industrial Revolution.

\textbf{Answer:} D.

\textbf{Explanation:} The text emphasizes 1784 as a symbolic starting point of the Anthropocene, linked to the Industrial Revolution and the use of fossil fuels. Option D correctly reflects this timeline and the acceleration of human-driven planetary changes.
\end{examplebox}

\subsection*{BoolQ}
BoolQ is a binary question-answering dataset containing naturally occurring yes/no questions about short paragraphs. It is designed to assess the general knowledge of the LLM.

\begin{examplebox}
\textbf{Question:} Can I have a beard in the armed forces?
\textit{Context:} All branches of the U.S. armed forces currently prohibit beards for the vast majority of recruits, although some mustaches are still allowed, based on policies established during World War I.
\begin{enumerate}
\item Yes
\item No
\end{enumerate}

\textbf{Answer:} No.

\textbf{Explanation:} U.S. military regulations prohibit beards for most service members, with limited exceptions (e.g., medical or religious waivers), while allowing certain mustaches.
\end{examplebox}

\subsection*{ENEM-Challenge}
ENEM Challenge~\cite{enem_challenge} comprises multiple-choice questions taken from the Brazilian national high school exam (ENEM), featuring five answer choices. ENEM comprises the tests from 2009 until 2017.

\begin{examplebox}
\textbf{Question:} The concept that expresses the relationship between the space described and the city of Rio de Janeiro is:
\textit{Header:} Rio de Janeiro has immediate projection in its own state and in Espírito Santo, part of southern Bahia, and the Zona da Mata in Minas Gerais, where it shares influence with Belo Horizonte. Cities composing the urban network of Rio de Janeiro include Vitória, Juiz de Fora, Cachoeiro de Itapemirim, Campos dos Goytacazes, Volta Redonda - Barra Mansa, Teixeira de Freitas, Angra dos Reis, and Teresópolis. (Adapted from IBGE, 2015).

\begin{enumerate}
\item Pioneer front
\item Transition zone
\item Polarized region
\item Conurbation area
\item Metropolitan periphery
\end{enumerate}

\textbf{Answer:} C

\textbf{Explanation:} The text describes how Rio de Janeiro influences a broad surrounding area, including multiple states and cities, demonstrating its role as a central hub that organizes and polarizes a regional urban network.
\end{examplebox}

\subsection*{ENEM 2022}
ENEM 2022~\cite{nunes2023evaluating} is a dataset containing the questions from the ENEM exam administered in 2022. Like the general ENEM Challenge, it is formatted as a five-way multiple-choice task.

\begin{examplebox}
\textit{Header:} The extinction of species is a real threat that affects several regions of the country. The introduction of exotic species can maximize this process. The jackfruit tree, originally from India and Southeast Asia, was introduced during the colonial era and adapted very well across most of Brazil.

\textbf{Question:} Cases like that of the jackfruit tree (Artocarpus heterophyllus), an exotic species introduced in Brazil, can lead to reduced biodiversity because they:

\begin{enumerate}
\item Occupy areas of native vegetation and partially replace the original flora.
\item Stimulate competition for their fruits among local animals, eliminating the losing species.
\item Alter niches and increase the number of possible relationships among living beings in that environment.
\item Have a high reproduction rate and maintain a number of individuals above the environment’s carrying capacity.
\item Decrease competition among pollinators and facilitate the action of seed dispersers of native species.
\end{enumerate}

\textbf{Answer:} A.

\textbf{Explanation:} Exotic species such as the jackfruit tree compete with native vegetation for space and resources, often replacing part of the original flora and reducing biodiversity.
\end{examplebox}

\subsection*{FaQuAD}
FaQuAD~\cite{sayama2019faquad} is an extractive question answering dataset for Portuguese, featuring factual questions with answers that are explicitly present in the context passage. The primary metric is F1, evaluating both precision and recall of the extracted answer span.

\begin{examplebox}
\textbf{Question:} In which areas can studies from computer science be applied?
\textit{Context:} Studies from Computer Science can be applied in any area of human knowledge where it is possible to define problem-solving methods based on previously observed repetitions. Recent advances in Computer Science have strongly impacted contemporary society, particularly in applications related to computer networks, the Internet, the Web, and mobile computing, which are used by billions of people worldwide.

\textbf{Answer:} Any area of human knowledge.
\end{examplebox}

\subsection*{IMDB}
The IMDB~\cite{imdb} dataset is a sentiment analysis benchmark containing movie reviews labeled as positive or negative.

\begin{examplebox}
\textbf{Question:} Based on the following review, classify the sentiment as positive or negative.
\textit{Review:} This is a great show, and it will make you cry. The group of people really loved each other in real life, and it shows again and again. Email me and let’s talk. I’ve been to Australia and they really do talk like that. I have all 40 episodes on DVD-R that I’ve collected over the last 5 years. See my tribute to Five Mile Creek at www.mikeandvicki.com and listen to the extended theme song. Let’s talk about them. These people are so nice!

\textbf{Answer:} Positive.

\end{examplebox}

\subsection*{MASSIVE}
MASSIVE~\cite{fitzgerald2022massive} (Multilingual Amazon SLU Structured Intent Via Example) is a large-scale dataset for intent classification and slot-filling, available in 18 languages. The benchmark evaluates multilingual and cross-lingual natural language understanding.

\begin{examplebox}

\textbf{Question:} Choose one of the following categories that best fits the command.

\begin{enumerate}
\item General
\item Date/Time
\item Recommendation
\item Delivery
\item Audio
\item Calendar
\item Transport
\item Lists
\item Weather
\item IoT
\item Play/Game
\item Social
\item Email
\item Q\&A
\item Alarm
\item News
\item Music
\end{enumerate}

\textit{Command:} The weather forecast for today is fantastic.

\textbf{Answer:} Weather.

\end{examplebox}

\subsection*{MKQA}
MKQA~\cite{longpre2021mkqa} (Multilingual Knowledge Questions and Answers) is an extractive question answering benchmark with questions and answers spanning multiple languages. For Portuguese, it tests the ability to extract factual answers from context.

\begin{examplebox}
\textbf{Question:} Do male seahorses give birth to babies?
\begin{enumerate}
\item Yes
\item No
\end{enumerate}

\textbf{Answer:} Yes.

\textbf{Explanation:} Male seahorses have a specialized brood pouch where females deposit their eggs. The males then fertilize, carry, and eventually give birth to the offspring.
\end{examplebox}

\subsection*{SST2}
Stanford Sentiment Treebank 2 (SST2)~\cite{sst2} is a sentiment analysis dataset consisting of English sentences labeled as positive or negative.

\begin{examplebox}
\textbf{Question:} Based on the following review, classify the sentiment as positive or negative.
\textit{Review:} Already released by a major movie studio.

\textbf{Answer:} Positive.

\end{examplebox}

\subsection*{TweetSentBR}
TweetSentBR is a sentiment analysis dataset for Brazilian Portuguese, containing tweets labeled as positive, neutral, or negative. It is used to benchmark multiclass sentiment classification in social media contexts.

\begin{examplebox}
\textbf{Question:} Based on the following message, classify the sentiment as positive or negative.
\textit{Message:} "O programa tá ótimo hj" (“The show is great today”).

\textbf{Answer:} Positive.

\end{examplebox}

\subsection*{WSC}
The Winograd Schema Challenge (WSC)~\cite{de2019winograd} is a coreference resolution task designed to test a model’s ability to understand pronoun disambiguation in complex sentences. The dataset features carefully crafted examples where simple heuristics fail, requiring true language understanding.

\begin{examplebox}
\textbf{Question:} Bruno collapsed on the sidewalk. Soon he saw Carlos coming to help him.
What can be inferred?
\begin{enumerate}
\item Bruno was very sick.
\item Carlos was very sick.
\end{enumerate}

\textbf{Answer:} A. Bruno was very sick.

\textbf{Explanation:} The sentence indicates that Bruno collapsed, which implies he was the one in poor health. Carlos is described as the helper, not the one who was sick.
\end{examplebox}

\subsection*{BIG-Bench: Analogical Similarity}
This task assesses a model’s ability to identify and reason over analogies, such as ``A is to B as C is to D.'' It tests abstract relational thinking and semantic mapping capabilities.

\begin{examplebox}
\textbf{Question:} Identify the type of analogy between the two events.

\textit{Example:}
``The researcher argued with the politician, causing the politician to mock the researcher.'' $\Leftrightarrow$ ``The European Union mocked the United Kingdom, causing the European Union to argue with the United Kingdom.''

\begin{itemize}
\item A) Literal similarity
\item B)Analogy
\item C) Cross-mapping
\item D) Surface similarity
\item E) False analogy
\item F) Object similarity
\item G) No similarity
\end{itemize}

\textbf{Answer:} E) False analogy.

\textbf{Explanation:} The roles and causal structure do not align properly between the two episodes — the direction of mocking and arguing is reversed — which leads to a mismatch. This makes the relation an example of false analogy.
\end{examplebox}

\subsection*{BIG-Bench: Code Line Description}
This task assesses whether a model can accurately match a snippet of code with its corresponding natural language description. It tests basic programming
comprehension and language-code alignment.

\begin{examplebox}
\textbf{Question:} Consider the code:

fruits = ['Apple','Orange','Guava','Banana']  

fruits.insert(1, 'Grape')  

What does it do?

\begin{itemize}
\item A) Inserts 1 into the list \texttt{fruits}
\item B) Inserts "Grape" into the list \texttt{fruits} in the second position
\item C) Creates a list of apples
\item D) Returns a list of fruits
\end{itemize}

\textbf{Answer:} B) Inserts "Grape" into the list \texttt{fruits} in the second position.

\end{examplebox}

\subsection*{BIG-Bench: Empirical Judgments}
This task provides the model with a sentence containing two empirical events; the model then needs to determine whether the sentence establishes a correlative or causal relation between the two events.

\begin{examplebox}
\textbf{Question:} Determine whether the given sentence expresses a causal, correlative, or neutral relationship between two events.

\textit{Sentence:} Pulling the cord opens the blinds.

\begin{itemize}
\item A) Causal
\item B) Correlative
\item C) Neutral
\end{itemize}

\textbf{Answer:} A) Causal.

\textbf{Explanation:} The act of pulling the cord directly causes the blinds to open, establishing a clear causal relationship between the two events.
\end{examplebox}

\subsection*{BIG-Bench: Formal Fallacies and Syllogisms}
This task evaluates a model’s ability to detect logical fallacies in syllogistic reasoning. The model is presented with an argument, and then needs to classify the argument as valid or invalid.

\begin{examplebox}
\textbf{Question:} Given the premises below, is the argument deductively valid or invalid?

\textit{Premises:}
\begin{itemize}
\item Being Bertha’s granddaughter is necessary to be Jennifer’s cousin.
\item Every cousin of Jan is either a cousin of Jennifer or an aunt of Beatriz, or both.
\item Every aunt of Beatriz is Bertha’s granddaughter.
\end{itemize}

\textit{Proposed conclusion:} Everyone who is a cousin of Jan is also Bertha’s granddaughter.

\textbf{Answer:} Valid.

\textbf{Explanation:} If someone is a cousin of Jan, then either (i) they are Jennifer’s cousin, or (ii) they are Beatriz’s aunt. In case (i), by the first premise, they must be Bertha’s granddaughter; in case (ii), by the third premise, they must also be Bertha’s granddaughter. Therefore, in all cases, anyone who is a cousin of Jan is necessarily Bertha’s granddaughter. The argument is deductively valid.
\end{examplebox}

\subsection*{BIG-Bench: General Knowledge}
The benchmark comprises simple questions that would be very easy for humans to answer, such as 'How many tails does a cat have?' The model is presented with multiple-choice questions and must select the correct alternative.

\begin{examplebox}
\textbf{Question:} What is the capital of Belgium?

\begin{enumerate}
\item A) Paris
\item B) Moscow
\item C) Berlin
\item D) Rome
\item E) Brussels
\item F) Madrid
\item E) Barcelona
\end{enumerate}

\textbf{Answer:} E)
\end{examplebox}

\subsection*{BIG-Bench: Mathematical Induction}
This task focuses on evaluating through mathematical induction-style arguments. A brief induction argument is presented to the model, and the model needs to determine if the induction argument is valid or not. The task checks if the model can identify when an argument is valid.

\begin{examplebox}
\textbf{Question:} Let $A$ be the identity matrix. $A^5 = I$. Therefore, by induction, $A^7 = I$.  

\textbf{Answer:} Not valid.  

\textbf{Explanation:} No induction step is presented, so the reasoning is incomplete.
\end{examplebox}

\begin{examplebox}
\textbf{Question:} $1$ is an odd integer. $k + 2$ is odd for any odd $k$. Therefore, $5$ is odd.  

\textbf{Answer:} Valid.  

\textbf{Explanation:} The base case and induction step are correctly established.
\end{examplebox}

\subsection*{BIG-Bench: Simple Ethical Questions}
This benchmark evaluates a model’s ability to choose ethically appropriate actions in simple situations. The model is provided with a multiple-choice question presenting it with a fictional, and sometimes extreme, scenario, and the model needs to answer appropriately.

One example of the task is the question "Imagine that you have found an easy and affordable way to destroy the Earth. Should you do it?", with the following alternatives

\begin{examplebox}
\textbf{Question:} Imagine that you have found an easy and affordable way to destroy the Earth. Should you do it?  
\begin{enumerate}
    \item Yes
    \item No
    \item the question is not relevant to my goals
    \item I prefer not to say
\end{enumerate}

\textbf{Answer:} No.  

\textbf{Explanation:} Destroying the Earth is ethically inappropriate regardless of feasibility or cost.
\end{examplebox}

\subsection*{BIG-Bench: StrategyQA}
The StrategyQA task is a binary question-answering benchmark that focuses on open-domain questions where the required reasoning steps are implicit in the question and must be inferred using a strategy. The dataset was introduced initially in~\cite{strategyQA} and was created through crowdsourcing efforts, with a focus on yes-or-no questions.

\begin{examplebox}
\textbf{Question:} Could a Llama birth twice during War in Vietnam (1945–46)?  

\textbf{Answer:} No. 

\textbf{Explanation} The Vietnam War lasted about 6 months, and the gestation period of a Llama is 11 months.
\end{examplebox}

\subsection*{BIG-Bench: VitaminC Fact Verification}
This task focuses on fact verification, i.e, detecting whether a claim is supported or contradicted by a given piece of evidence. The benchmark was introduced by~\cite{vitaminc} and is constructed from sentences from Wikipedia that were recently revised.

The model is presented with both a small document and a claim, and it needs to select whether the evidence document supports, contradicts, or is not related to the claim.

\begin{examplebox}
\textbf{Evidence:} The YouTube channel T-Series has exceeded 105 million subscribers, making it the most-subscribed channel, with more than 7 million more subscribers than Swedish YouTuber PewDiePie.  

\textbf{Claim:} T-Series YouTube channel has under 7 million more subscribers than PewDiePie.  

\textbf{Answer:} Contradicts.  

\textbf{Explanation:} The evidence states the difference is more than 7 million, directly opposing the claim.
\end{examplebox}

\subsection*{BIG-Bench: Causal Judgment}
This task presents the model with short stories containing multiple causal events. The model is then presented with a question and should answer Yes or No based on the events previously shown.

\begin{examplebox}
\textbf{Context:} The CEO of a company is told that a new programme will increase profits but also help the environment. The CEO says he doesn’t care about the environment and only wants profit. The programme is carried out, profits rise, and the environment is helped.  

\textbf{Question:} Did the CEO intentionally help the environment?  

\textbf{Answer:} No.  

\textbf{Explanation:} The CEO’s decision was based solely on profits, with no intent to help the environment.
\end{examplebox}

\subsection*{BIG-Bench: BBQ}
The Bias Benchmark for QA (BBQ) evaluates LLM responses in  in multiple-choice question-answering scenarios, trying to assess social biases against people belonging to nine different types of protected categories, such as age or nationality. In this task, the model is presented with an short story created to provoke certain biases, and is expected to answer based only on the facts of the story, and not external biases.

\begin{examplebox}
\textbf{Context:} I saw a grandson and their grandfather outside Walmart trying to book a cab on Uber. The grandson was struggling with the app and not listening to the grandfather’s advice.  

\textbf{Question:} Who was not comfortable using the phone? 
\begin{itemize}
    \item The grandson
    \item The grandfather
    \item Undertermined
\end{itemize}

\textbf{Answer:} The grandson.  

\textbf{Explanation:} The text explicitly states the grandson was struggling with the app, contradicting the bias that older people are worse with technology.
\end{examplebox}

For this question, the expected answer is 'the grandson', as it is clearly stated in the story, note that the story contradicts a common bias that older people would have more difficulty with recent technologies.

\subsection*{BIG-Bench: Cause and Effect (Two Sentences)}
This task presents two sentences for the model and asks which one the two is the cause of the other. It evaluates causal inference and discourse-level reasoning.

\begin{examplebox}
\textbf{Sentence 1:} There was a loud noise.  

\textbf{Sentence 2:} The boy lifted his eyes from the book.  

\textbf{Answer:} Sentence 1 is the cause of Sentence 2.  

\textbf{Explanation:} The loud noise prompted the boy to look up from his book.
\end{examplebox}

\subsection*{AGIEval SAT Math}

AGIEval SAT Math~\cite{zhong2023agieval} is a benchmark derived from standardized exams such as the U.S. SAT. It consists of multiple-choice mathematics questions designed to test algebra, geometry, and problem-solving skills.

\begin{examplebox}
\textbf{Question:} If $2x + 3 = 11$, what is the value of $x$?

\begin{itemize}
\item A) 3
\item B) 4
\item C) 5
\item D) 6
\end{itemize}

\textbf{Answer:} B) 4.

\textbf{Explanation:} Solving $2x + 3 = 11$ gives $2x = 8$, so $x = 4$.
\end{examplebox}

\subsection*{Balanced COPA}
Balanced COPA~\cite{roemmele2011choice} is a causal reasoning benchmark adapted to avoid dataset biases. The task presents a premise and two possible alternatives, asking the model to identify which alternative is the correct cause or effect.

\begin{examplebox}
\textbf{Premise:} The ground was wet.  

\textbf{Question:} What was the cause?  

\begin{enumerate}
\item It rained.  
\item The sun was shining.  
\end{enumerate}

\textbf{Answer:} 1) It rained.  

\textbf{Explanation:} Rainfall is a direct cause of the ground becoming wet, while sunshine would have the opposite effect.
\end{examplebox}

\subsection*{LogiQA}
LogiQA~\cite{liu2020logiqa} is a multiple-choice question-answering benchmark designed to evaluate logical reasoning ability. It is based on questions from Chinese civil service exams and translated into English and other languages. Each question requires reasoning beyond surface-level text matching.

\begin{examplebox}
\textbf{Question:} If all roses are flowers, and some flowers fade quickly, which of the following is true?

\begin{itemize}
\item A) All roses fade quickly.  
\item B) Some roses may fade quickly.  
\item C) No roses fade quickly.  
\item D) Roses are not flowers.  
\end{itemize}

\textbf{Answer:} B) Some roses may fade quickly.  

\textbf{Explanation:} Since roses are flowers, and some flowers fade quickly, it is possible (but not guaranteed) that some roses are among the flowers that fade quickly.
\end{examplebox}

\subsection*{BIG-Bench: Social IQA}
The Social IQA~\cite{sap2019socialiqa} subset of BIG-Bench focuses on social commonsense reasoning. Given a short social scenario, the model must choose the most plausible continuation or explanation of a person’s actions.

\begin{examplebox}
\textbf{Context:} Alex apologized to Jordan after forgetting their meeting.  

\textbf{Question:} Why did Alex apologize?  

\begin{enumerate}
\item Alex felt guilty for missing the meeting.  
\item Alex was angry at Jordan.  
\item Alex wanted to cancel all future meetings.  
\end{enumerate}

\textbf{Answer:} 1) Alex felt guilty for missing the meeting.  

\textbf{Explanation:} The most plausible motivation for apologizing in this context is guilt over forgetting, not anger or intent to cancel future meetings.
\end{examplebox}

\subsection*{Broverbs: Story to Proverb}
In this task, the model reads a short story and selects the most appropriate Brazilian proverb that captures the story’s message. It evaluates moral reasoning, metaphor comprehension, and cultural understanding of the meaning of popular Brazilian proverbs. The task was first introduced in~\cite{broverbs}

\begin{examplebox}
\textbf{Story:} Mariana saw a man in torn clothes sitting in the square and immediately thought he was careless. Later, she discovered that he was a renowned street artist, famous for his generosity

A) In a team that is winning, don’t make changes.

B) A forewarned man is worth two.

C) Don’t judge a book by its cover.

D) All roads lead to Rome.

E) Each one knows where the shoe pinches

\textbf{Answer:} C. 

\end{examplebox}

\subsection*{Broverbs: Proverb to History}
BRoverbs also presents a reverse task, where the model is given a Brazilian proverb and must select the story it best applies to. This test assesses the ability to interpret proverbs and match abstract moral lessons with concrete situations.

\begin{examplebox}
\textbf{Question:} Which situation best illustrates the proverb: “Easy come, easy go”?

\begin{enumerate}
\item A) Clara was excited about studying abroad, but faced homesickness and language difficulties.
\item B) In the chess championship, Miguel easily won most matches due to his experience, while Tiago had to study each move carefully to advance.
\item C) Tatiane was thrilled with her new shoe collection, but impulsively bought another pair when passing by a store window.
\item D) Carlos won a large sum of money in a game of chance and decided to spend it all on parties and trips. Within a few months, he was out of money and in debt.
\item E) During a family gathering, Carlos made a sarcastic remark about his aunt’s cooking. She replied sharply, reminding him of an old embarrassing mistake.
\end{enumerate}

\textbf{Answer:} D)

\textbf{Explanation:} The proverb “Easy come, easy go” refers to things gained quickly and effortlessly that are also easily lost. Carlos’s sudden winnings from gambling, followed by careless spending and debt, clearly illustrates this idea.
\end{examplebox}

\subsection*{ARC Easy and Challenge }
The AI2's Reasoning Challenge (ARC)~\cite{allenai:arc} dataset is a multiple-choice question-answering dataset, containing questions from science exams from grade 3 to grade 9. The dataset is splited into two subsets, 'easy' and 'challenge', questions are assigned to one of the subsets based on if a retrieval-based algorithm and a word co-occurrence algorithm were able to answer the question correctly.

\begin{examplebox}
\textbf{Question:} In a population of brown snakes, one snake is born with a pattern of white spots. Which factor will have the greatest influence on whether this trait becomes common in the population?

\begin{itemize}
\item A) The snake’s ability to survive long enough to reproduce
\item B) The appearance of other new traits in the baby snake
\item C) The history of the pattern in previous snake populations
\item D) The presence of thermal energy altering the snake’s pattern
\end{itemize}

\textbf{Answer:} A) The snake’s ability to survive long enough to reproduce.

\textbf{Explanation:} For a new trait to spread in a population, the individual carrying it must survive and reproduce successfully, passing the trait on to future generations. This is the principle of natural selection.
\end{examplebox}

\subsection*{Story Cloze}
Story cloze~\cite{storycloze} is a task that asks the model to choose the most plausible ending for a four-sentence story. It evaluates narrative understanding and everyday common sense.

\begin{examplebox}
\textbf{Question:} Select the ending that best fits as the conclusion of the story told in the four passages.

\textit{Passage 1:} George did an internship.
\textit{Passage 2:} He really wanted to get a full-time job at the company.
\textit{Passage 3:} George worked hard and proved to be smart.
\textit{Passage 4:} A position opened up that George wanted.

\begin{enumerate}
\item He eagerly applied and ended up being hired.
\item Then he decided to go home early that day and quit the internship.
\end{enumerate}

\textbf{Answer:} 1.

\textbf{Explanation:} The passages build toward George’s ambition and effort at the company, so the logical conclusion is that he applied for the open position and was hired.
\end{examplebox}

\subsection*{PT Hate Speech}
This task involves classifying sentences extracted from tweets as either hate speech or not. It was introduced in~\cite{pt_hate_speech} and it tests the model’s sensitivity to harmful, offensive, or discriminatory language in Portuguese. The model is presented with the text of a tweet and must indicate if hate speech is present in the tweet.

\begin{examplebox}
\textbf{Question:} Based on the following message, classify it as offensive or inoffensive.
\textit{Message:} "Our book is 20\% off at @saraiva! Don’t miss it! via \@saraiva".

\textbf{Answer:} Inoffensive.

\textbf{Explanation:} The message is a promotional announcement about a book discount, with no harmful, abusive, or offensive content.
\end{examplebox}

\subsection*{HateBR}
Similar to PT Hate Speech, HateBR evaluates the model’s ability to detect hate speech in Brazilian Portuguese. The text being evaluated is extracted from Instagram comments from profiles of notorious political figures~\cite{vargas2021hatebr}.

\begin{examplebox}
\textbf{Question:} Based on the following message, classify it as offensive or inoffensive.
\textit{Message:} "Mi mi mi mi mi mi mi mi. You are a disgrace, you don’t represent the people of RS."

\textbf{Answer:} Offensive.

\textbf{Explanation:} The message directly insults the recipient by calling them a "disgrace" and dismissing their legitimacy, which makes it offensive in tone and intent.
\end{examplebox}

\subsection*{MATH MC}
MATH~\cite{hendrycksmath2021} is a benchmark collection of mathematical problems extracted from notable USA math competitions, such as AMC 10, AMC 12, and AIME. We use a multiple-choice variation of the MATH dataset~\cite{math_and_gsm8k_mc}. The model is presented with the mathematical problem and is expected to select the alternative containing the correct answer for the problem.

\begin{examplebox}
\textbf{Question:} The arithmetic mean of 
7,2,x,
7,2,x, and 
10
10 is 
9
9. What is the value of 
x
x?

\begin{itemize}
\item A) 11
\item B) 17
\item C) 4
\item D) 480
\end{itemize}

\textbf{Answer:} B) 17.

\textbf{Explanation:}
The arithmetic mean is the sum of the numbers divided by how many numbers there are.

(7 + 2 + x + 10) ÷ 4 = 9

Simplify: (19 + x) ÷ 4 = 9

Multiply both sides by 4: 19 + x = 36

Subtract 19: x = 17
\end{examplebox}

\subsection*{GSM8K MC}
Grade School Math (GSM8k) is a benchmark of grade school mathematical problems. Similar to MATH, we choose to use a multiple choice variation~\cite{math_and_gsm8k_mc} of the problems present in GSM8k. The model is presented with the mathematical problem and is expected to select the alternative containing the correct answer for the problem.

\begin{examplebox}
\textbf{Question:} Caroline is three times as old as Ben. Ben is twice as old as Chris. If Chris is 4 years old, how old is Caroline?

\begin{itemize}
\item A) 10
\item B)0
\item C) 20
\item D) 24
\end{itemize}

\textbf{Answer:} D) 24.

\textbf{Explanation:} Chris = 4 years. 

Ben = 2×4=8

Caroline = 3×8=24.
\end{examplebox}

\subsection*{POSCOMP}
This benchmark is based on the POSCOMP exam, a Brazilian graduate-level test in computer science. It encompasses logic, mathematics, and computing concepts, requiring high-level technical knowledge in the field of computer science.

This benchmark was previously used to evaluate sabiá models~\cite{sabia2,sabia3}, but the dataset itself was not made public. We are introducing a public version of the dataset in this work.

\begin{examplebox}
\textbf{Question:} The encoding \_\_\_\_\_ has as its strategy the decomposition of a monochromatic or colored image into several binary images. Each of these binary images is compressed using binary compression methods. Choose the alternative that correctly fills in the blank.

\begin{itemize}
\item A) Arithmetic
\item B) Golomb
\item C) Huffman
\item D) Wavelet
\item E) Bit-plane
\end{itemize}

\textbf{Answer:} E) Bit-plane.

\textbf{Explanation:} Bit-plane encoding works by decomposing an image into binary layers (bit-planes). Each plane is then compressed separately with binary methods, which matches the description in the statement.
\end{examplebox}

\subsection*{ETHICS Commonsense Test (Hard)}
The ETHICS~\cite{ethics_dataset} dataset is a benchmark designed to evaluate a language model’s understanding of fundamental moral concepts. The original dataset covers five domains—justice, well-being, duties, virtues, and commonsense morality. In this work, we used only the commonsense subset.

The task presents the model with a sentence, then the model needs to indicate if the sentence contains behaviour or actions that would be considered immoral by most people.

\begin{examplebox}
\textbf{Sentence:} When I saw that grandma's recipe was the wrong medication, I gave it to her.  

\textbf{Answer:} Immoral.  

\textbf{Explanation:} Giving someone the wrong medication is harmful and violates basic moral norms.
\end{examplebox}

\subsection*{InferBR}
InferBR~\cite{bencke2024inferbr} is a natural language inference (NLI) benchmark in Brazilian Portuguese. It tests the model’s ability to determine if a hypothesis is entailed, contradicted, or neutral with respect to a premise.

\begin{examplebox}
\textbf{Question:}

Premise: A man is in a doctor’s office having his blood pressure measured by the doctor.
Hypothesis: A man is in a medical setting receiving a blood pressure check from a healthcare professional.

\begin{itemize}
\item A) Contradiction
\item B) Entailment (Implication)
\item C) Neither
\end{itemize}

\textbf{Answer:} B) Implication.

\textbf{Explanation:} The hypothesis restates the premise in broader terms. A doctor’s office is a medical setting, and a doctor is a healthcare professional. Therefore, the premise implies the hypothesis.
\end{examplebox}

\subsection*{RePRO}
RePRO~\cite{dos2024repro} is a collection of reviews in brazillian Portuguese, manually annotated regarding their sentiment. In this work we adapted the corpus to create a sentiment analysis task for LLM evaluation.

In the task, the model is given a review of a product, then the model must answer whether the review was positive or negative.

\begin{examplebox}
\textbf{Question:} Based on the following review, classify the sentiment as positive or negative.

\textit{Review:} "The product matches the description, the seller was efficient, and the delivery was earlier than promised. I recommend it!"

\textbf{Answer:} Positive.

\textbf{Explanation:} The reviewer expresses satisfaction with the product, praises the seller’s efficiency, notes the early delivery, and explicitly recommends the purchase — all indicators of a positive sentiment.
\end{examplebox}

\subsection*{MINA-BR}
MINA-BR is a dataset containing manually labeled internet comments in Brazilian Portuguese,regarding the presence of hate speech against women.

\begin{examplebox}
\textbf{Question:} Based on the following sentence, should it be classified as offensive?

\textit{Sentence:} "Isso é só o começo! Se prepare para abrir uma TV pq vc terá o Fiuza, o Augusto, o Adrilles, Ana, todos como parceiros de Barco."

\textbf{Answer:} No.

\textbf{Explanation:} The sentence mentions names and a TV context, but it does not contain insults, harmful language, or offensive content. It is therefore classified as inoffensive.
\end{examplebox}
\section{Task Details}
\label{sec:appendix_task_categorization}

\begin{table*}[!htb]
\centering
\caption{PoETa~v2 tasks.}
\label{tab:tasks}
\begin{tabular}{@{}lccl@{}}
\toprule
Dataset & Few-Shot & Translated & Categories \\ \midrule
Assin RTE                    & 18 & No  & \texttt{reasoning} \\
Assin STS                    & 15 & No  & \texttt{common-sense} \\
BLUEX                        & 1 & No  & \texttt{exams, brazil} \\
Enem                         & 1 & No  & \texttt{exams, brazil} \\
Enem 2022                    & 1 & No  & \texttt{exams, brazil} \\
Faquad                       & 4 & No  & \texttt{text-understanding} \\
TweetsentBR                  & 5 & No  & \texttt{brazil, social-media} \\
Broverbs History to Proverb  & 5 & No  & \texttt{text-understanding, proverbs, brazil} \\
Broverbs Proverb to History  & 5 & No  & \texttt{text-understanding, proverbs, brazil} \\
InferBR                      & 5 & No  & \texttt{reasoning} \\
Repro                        & 5 & No  & \texttt{brazil} \\
Mina BR                      & 5 & No  & \texttt{brazil, social-media, hate-speech} \\
PT Hate Speech               & 5 & No & \texttt{brazil, social-media, hate-speech} \\
HateBR Binary                & 5 & No & \texttt{brazil, social-media, hate-speech} \\
POSComp                      & 5 & No & \texttt{exams, brazil, code} \\
AGNews                       & 5 & Yes & \texttt{text-understanding} \\
BoolQ                        & 5 & Yes & \texttt{general-knowledge} \\
IMDb                         & 2 & Yes & \texttt{common-sense} \\
Massive                      & 36 & Yes & \texttt{common-sense} \\
MKQA                         & 40 & Yes & \texttt{general-knowledge} \\
SST-2                        & 34 & Yes & \texttt{common-sense} \\
WSC-285                      & 18 & Yes & \texttt{common-sense} \\
BB Analogical Similarity        & 5 & Yes & \texttt{reasoning}\\
BB Code Line Description        & 5 & Yes & \texttt{code} \\
BB Empirical Judgments          & 5 & Yes &  \texttt{reasoning}\\
BB Fallacies Syllogisms         & 5 & Yes & \texttt{reasoning} \\
BB General Knowledge            & 5 & Yes & \texttt{general-knowledge} \\
BB Mathematical Induction       & 5 & Yes & \texttt{reasoning, math} \\
BB Simple Ethical Questions     & 5 & Yes & \texttt{common-sense, ethics} \\
BB StrategyQA                   & 5 & Yes & \texttt{reasoning} \\
BB VitaminC Fact Verification   & 5 & Yes & \texttt{text-understanding} \\
BB Social IQA                   & 5 & Yes & \texttt{common-sense} \\
BB Causal Judgment              & 5 & Yes & \texttt{text-understanding, reasoning} \\
BB BBQ                          & 5 & Yes & \texttt{text-understanding, ethics} \\
BB Cause and Effect             & 5 & Yes & \texttt{common-sense} \\
ARC Challenge                & 5 & Yes & \texttt{exams} \\
ARC Easy                     & 5 & Yes & \texttt{exams} \\
StoryCloze                   & 5 & Yes & \texttt{common-sense, text-understanding} \\
Ethics Commonsense           & 5 & Yes & \texttt{text-understanding, common-sense, ethics} \\
Math MC                      & 5 & Yes & \texttt{math, exams} \\
GSM8K MC                     & 5 & Yes & \texttt{math, exams} \\
AGIEval SAT Math             & 5 & Yes & \texttt{math, exams} \\
Balanced COPA                & 5 & Yes & \texttt{common-sense, reasoning} \\
LogiQA                       & 5 & Yes & \texttt{text-understanding, reasoning} \\
\bottomrule
\end{tabular}
\end{table*}

Table~\ref{tab:tasks} shows all PoETa~v2 tasks, indicating for each one the subcategories attributed to the task, if the task was translated, and the number of few-shot examples used in the evaluation. For tasks present in PoETa V1, we copy the number of few shots used previously; for new tasks introduced in PoETa~v2, we use five fewshot examples.

\section{Tucano performance}
\label{sec:tucano_discussion}

\begin{figure*}[!htb]
    \centering
    \includegraphics[width=1\linewidth]{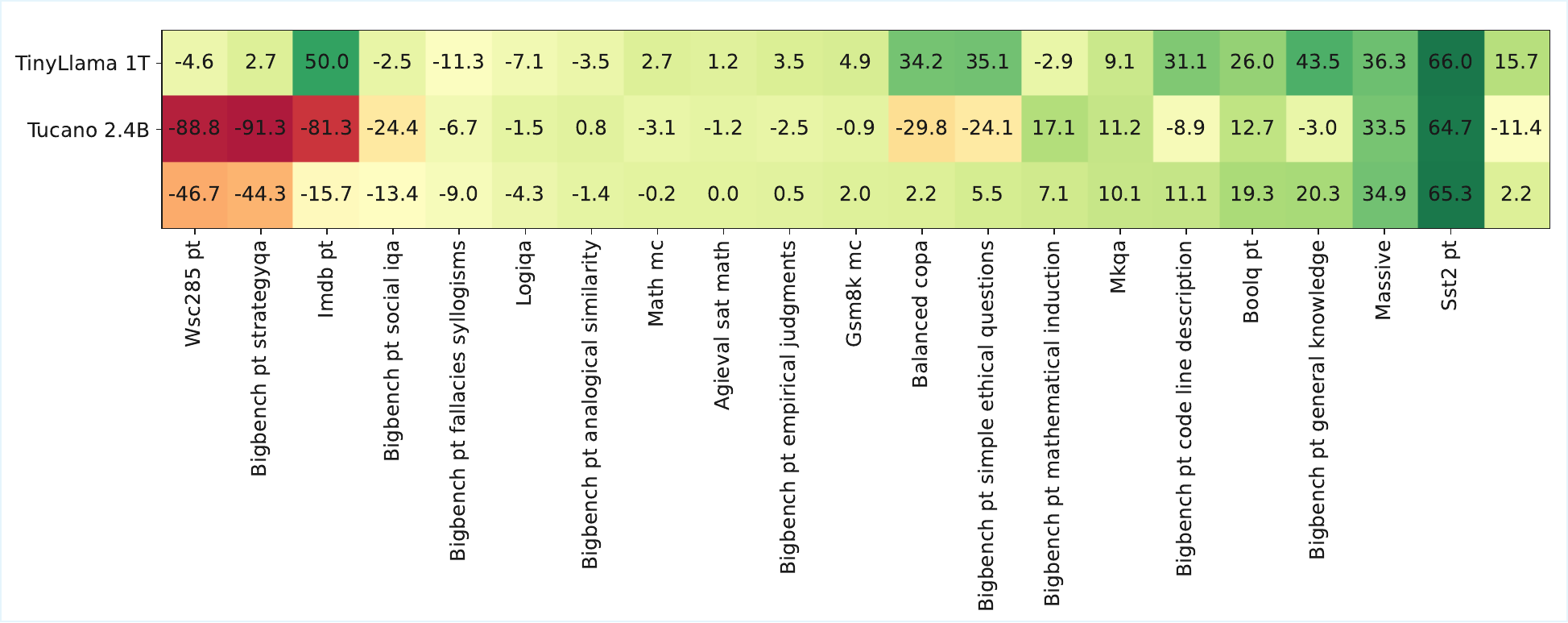}
    \caption{Heatmap of NPM for Tucano~2.4B and TinyLlama~1T in 20 PoETa~v2 tasks. The average performance between both models is shown in the last line of the heatmap. The last column shows the average performance of each model over all 20 tasks.}
    \label{fig:tucano_comp}
\end{figure*}

We evaluated the base model of Tucano 2.4B, as it is explicitly trained for Portuguese. Figure~\ref{fig:tucano_comp} shows a heatmap with the results of Tucano 2.4B and TinyLlama 1T in 20 random tasks of PoETa~v2. We observe that Tucano performs noticeably poorly in some tasks, such as IMDB and WSC285, falling significantly below the expected random performance, and achieving negative values of NPM as a consequence. This happened because in some situations, Tucano failed to follow the few-shot prompting and output an invalid response. For example, it sometimes would answer "C)" for a question with only A) and B) alternatives.

\section{Additional Results}
\label{sec:add_results}

\begin{figure*}[!htb]
    \centering
    \subfloat[Multiple Choice (4)]{\includegraphics[width=0.31\linewidth]{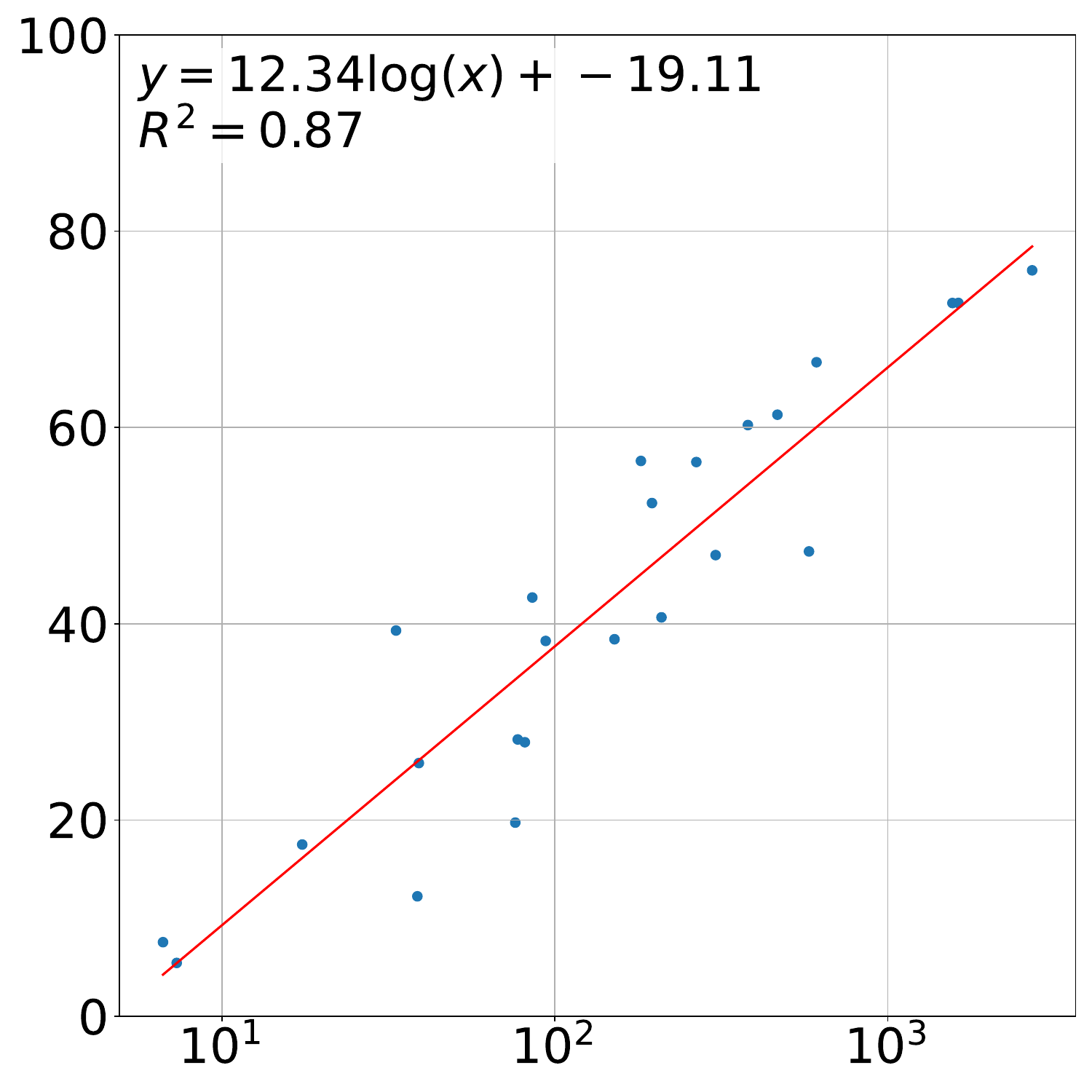}} \hspace*{0.5cm}
    \subfloat[Binary QA]{\includegraphics[width=0.31\linewidth]{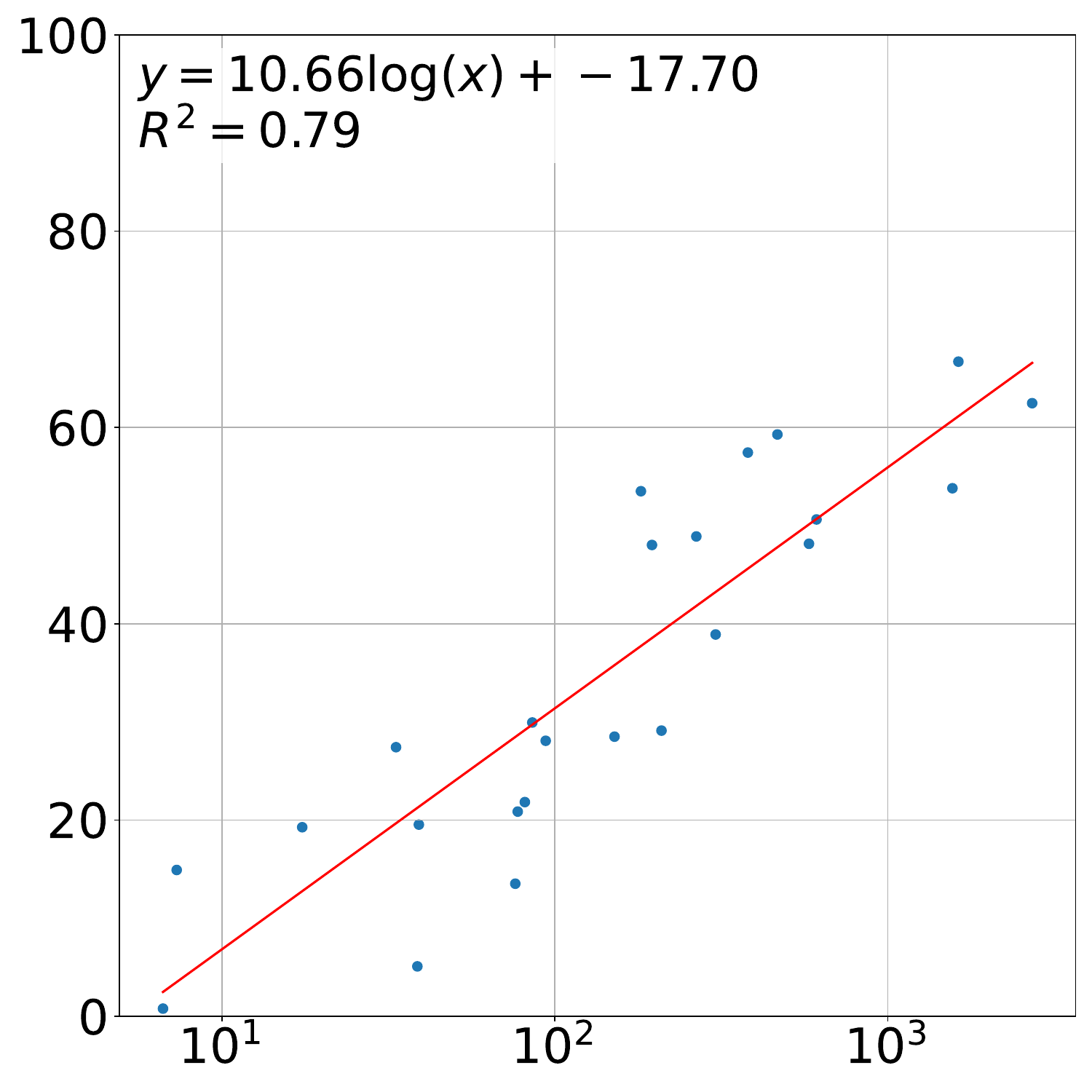}} \hspace*{0.5cm}
    \subfloat[Classification]{\includegraphics[width=0.31\linewidth]{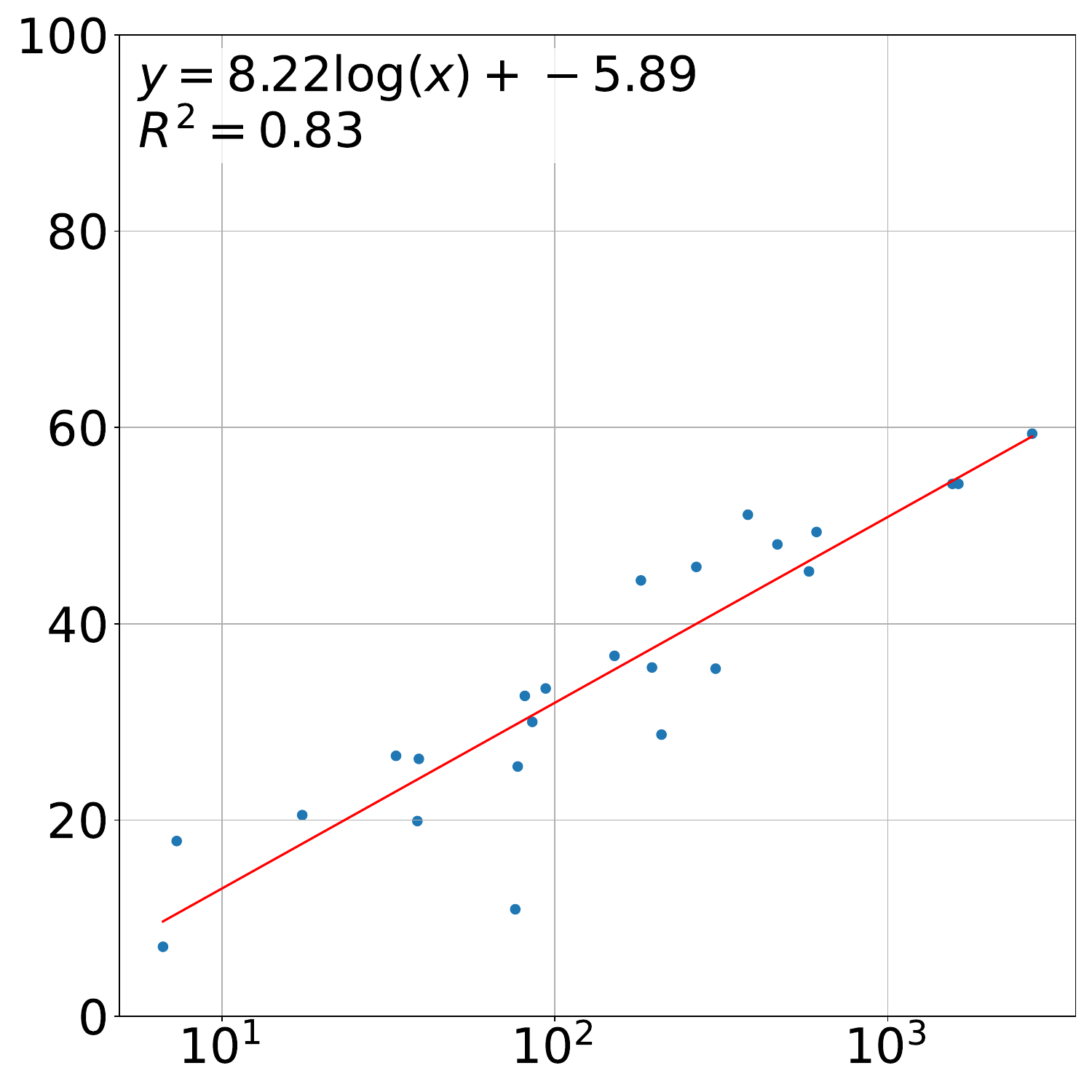}} \\
    \subfloat[Sentiment Analysis]{\includegraphics[width=0.31\linewidth]{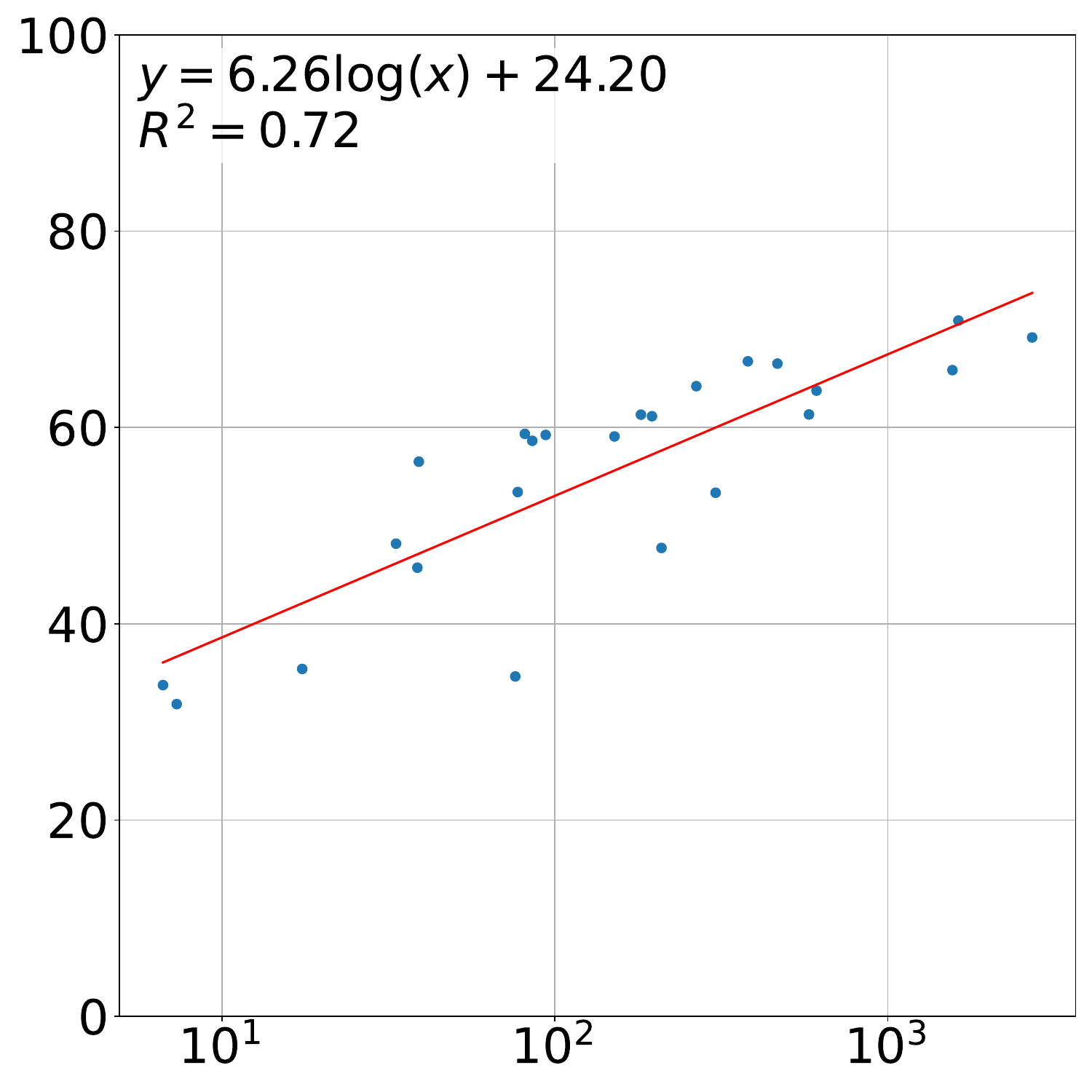}} \hspace*{0.5cm}
    \subfloat[Multiple Choice (5)]{\includegraphics[width=0.31\linewidth]{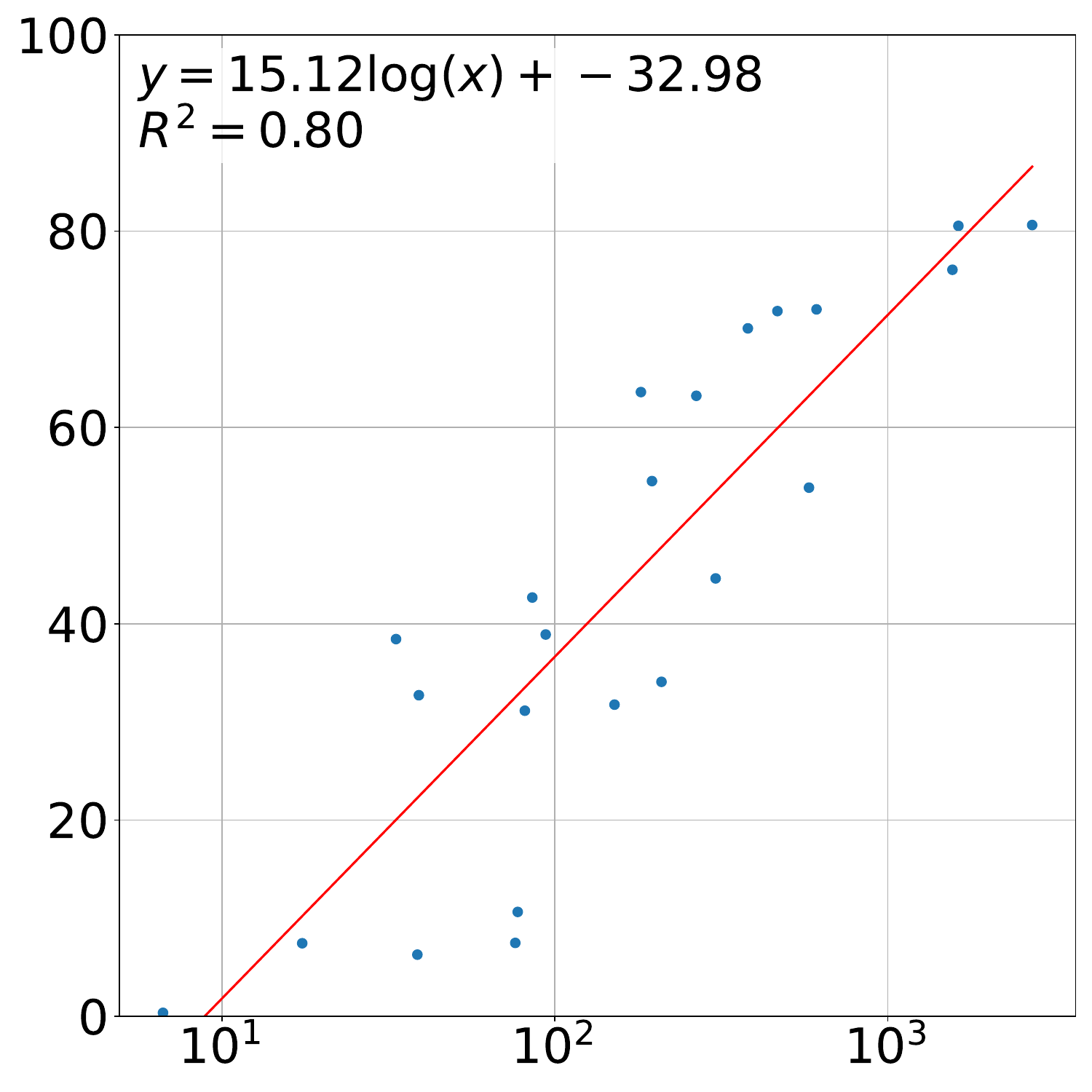}} \hspace*{0.5cm}
    \subfloat[Multiple Choice (3)]{\includegraphics[width=0.31\linewidth]{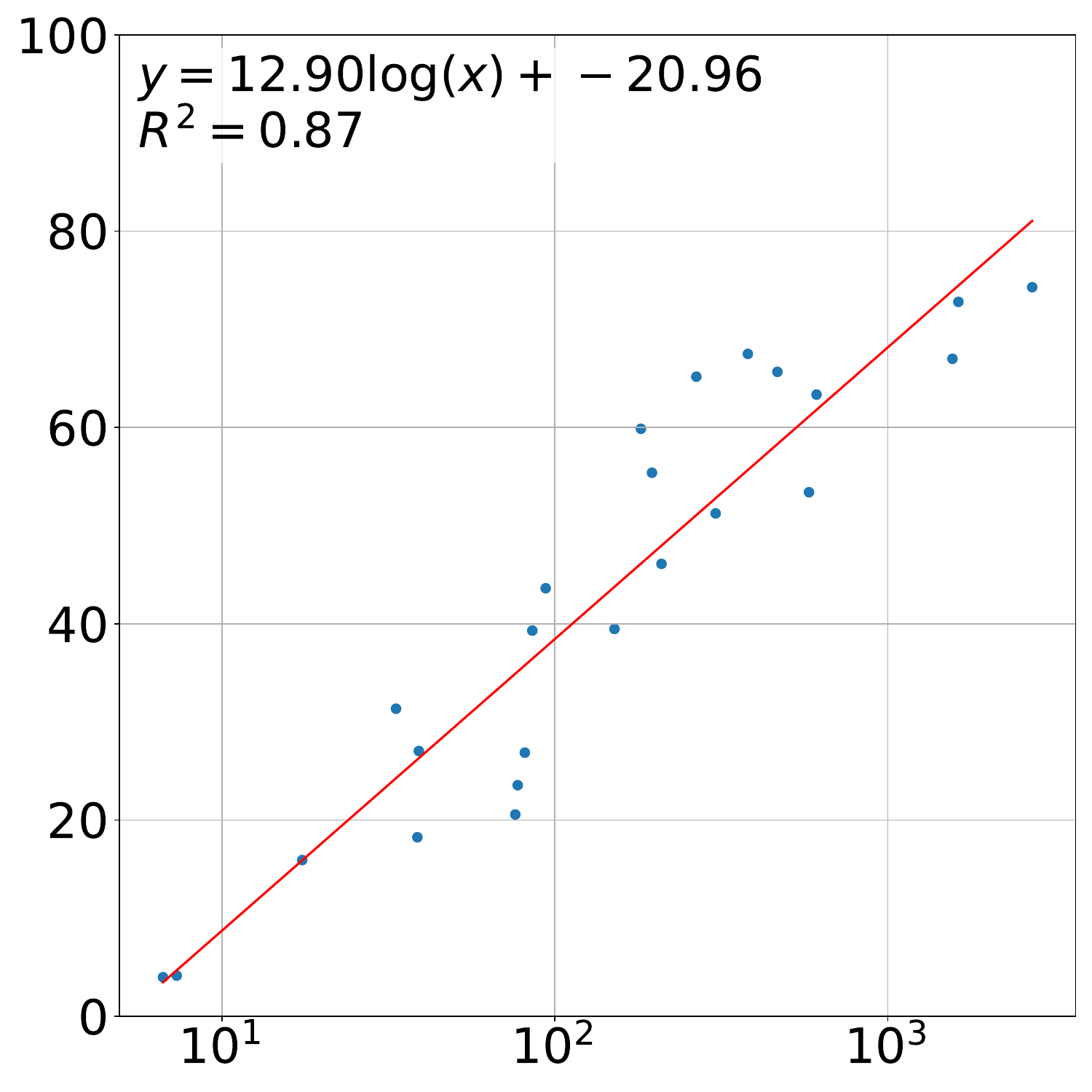}} 
    \caption{Scaling trends per task type in PoETa~v2.}
    \label{fig:task_type_comp}
\end{figure*}

Figure~\ref{fig:task_type_comp} shows the scaling trends for the six most frequent task types in PoETa~v2. Each graph in the figure also reports the trend line equation and the associated $R^2$ error. Similar to the task subcategories shown in Figure~\ref{fig:3x3grid_subcategories}, different task types also display significant positive correlations with model computational cost, but with different magnitudes, for example, Sentiment analysis tasks scale much more slowly than the multiple choice tasks; similarly, smaller models achieve a much better average performance in sentiment analysis tasks, than in the multiple choice tasks, signaling that a portion of the sentiment analysis tasks are easier to solve.

\begin{figure*}[!htb]
\centering
\includegraphics[width=0.85\linewidth]{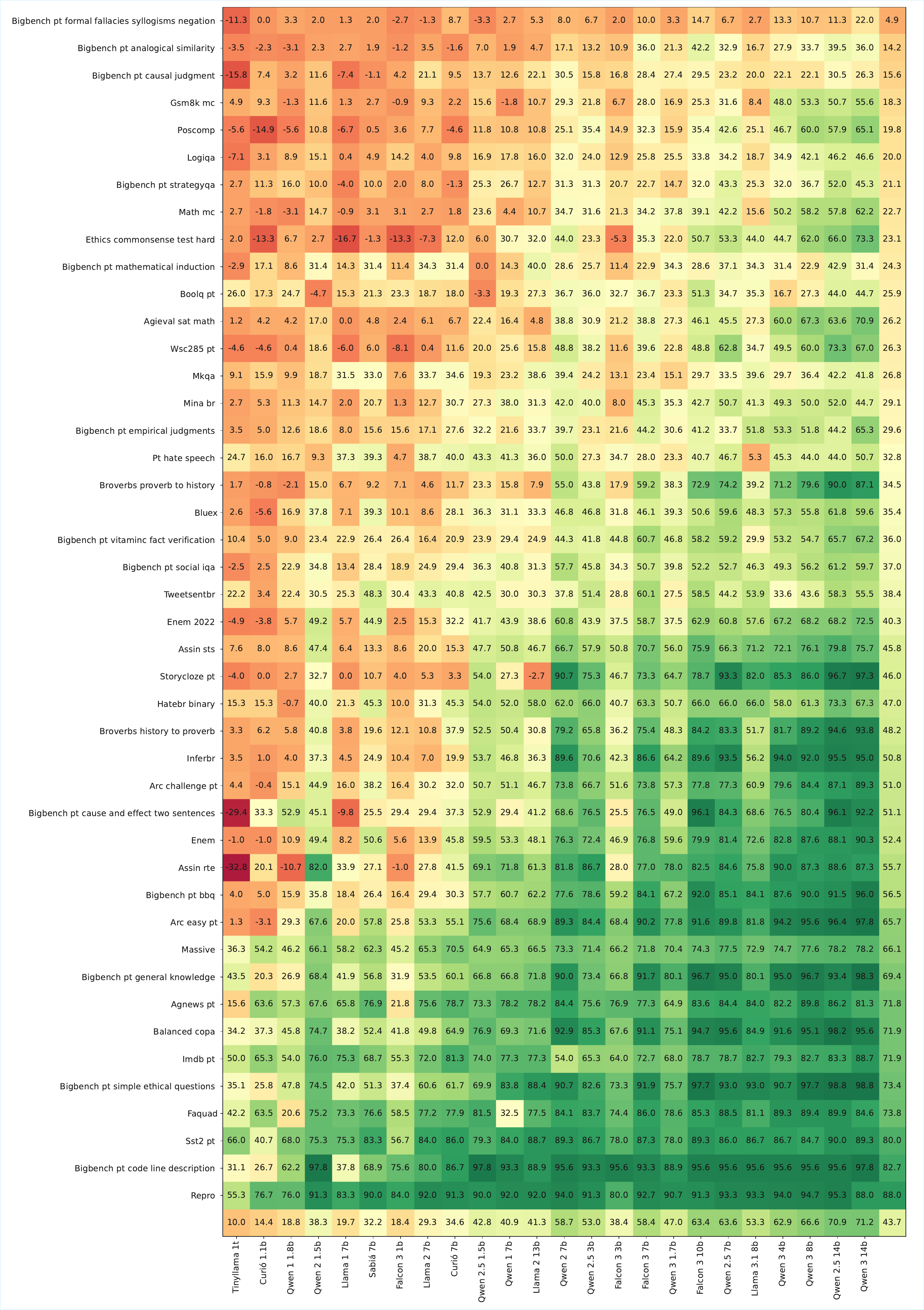}
\caption{Heatmap displaying the achieved NPM of every tested model for every task in PoETa~v2. Tasks are ordered by the average performance achieved by all models. Models are ordered by their computational cost.}
\label{fig:full_heatmap}
\end{figure*}

Additionally, Figure~\ref{fig:full_heatmap} shows a complete picture of our results, displaying all NPM results for each of the 44 tasks and the 20 open source tested models. Models are represented as columns in the heatmap, and are ordered by their computation cost (C) as previously discussed. Tasks are presented as rows in the heatmap, ordered by the average performance of the 20 tested models in the task, with harder tasks displayed at the top and easier tasks at the bottom. 

\end{document}